\documentclass[10pt,twocolumn,letterpaper]{article}

\usepackage{cvpr}              
\usepackage{booktabs}
\usepackage{multirow}
\usepackage{longtable}
\usepackage{colortbl,hhline}
\usepackage[accsupp]{axessibility}  
\usepackage{ulem}
\usepackage[dvipsnames]{xcolor}

\definecolor{cvprblue}{rgb}{0.21,0.49,0.74}
\usepackage[pagebackref,break links,color links,citecolor=cvprblue]{hyperref}
\usepackage{multirow}
\usepackage{array}
\usepackage[accsupp]{axessibility}  
\usepackage{tabu}
\DeclareMathOperator*{\argmax}{argmax}

\usepackage{csquotes}
\def\paperID{7001} 
\def\confName{CVPR}
\def\confYear{2024}
\title{CPLIP: Zero-Shot Learning for Histopathology with Comprehensive Vision-Language Alignment}
\vspace{-.6cm}
\author{Sajid Javed$^{1}$,~Arif Mahmood$^{2}$,~Iyyakutti Iyappan Ganapathi$^{1,*}$,~Fayaz Ali Dharejo$^{1}$,\\~Naoufel Werghi$^{1,*}$,~Mohammed Bennamoun$^{3}$\\$^{1}$Department of Computer Science, $^{*}$C2PS, Khalifa University of Science and Technology, UAE\\$^{2}$Information Technology University of the Punjab, Pakistan,~$^{3}$The University of the Western Australia\\\\ \textcolor{red}{\textbf{This paper has been accepted in CVPR 2024}}\\
}

\begin{document}
\vspace{-.6cm}
\maketitle
\vspace{-.6cm}
\begin{abstract}
This paper proposes Comprehensive Pathology Language Image Pre-training (CPLIP), a new unsupervised technique designed to enhance the alignment of images and text in histopathology for tasks such as classification and segmentation. This methodology enriches vision-language models by leveraging extensive data without needing ground truth annotations. CPLIP involves constructing a pathology-specific dictionary, generating textual descriptions for images using language models, and retrieving relevant images for each text snippet via a pre-trained model. 
The model is then fine-tuned using a many-to-many contrastive learning method to align complex interrelated concepts across both modalities.
Evaluated across multiple histopathology tasks, CPLIP shows notable improvements in zero-shot learning scenarios, outperforming existing methods in both interpretability and robustness and setting a higher benchmark for the application of vision-language models in the field.
To encourage further research and replication, the code for CPLIP is available on GitHub at \textcolor{magenta}{https://cplip.github.io/}
\end{abstract}

\vspace{-.6cm}
\section{Introduction}
\label{sec:intro}
Vision Language (VL) models have substantially progressed, enhancing a broad spectrum of vision applications with their ability to understand open vocabularies and demonstrate capabilities for zero-shot transfer \cite{zhang2023vision, zhang2021vinvl, du2022learning, zhou2022learning, li2023citetracker, shao2023prompting}. Key to this progress is the effective alignment of visual and linguistic data, done using large datasets with paired images and text \cite{radford2021learning}. The Contrastive Language-Image Pretraining (CLIP) model exemplifies this evolution, using contrastive learning to align visual and text embeddings on a large scale \cite{radford2021learning}.
%
%
%
\begin{figure}[t!]
\centering
\includegraphics[width=\linewidth]{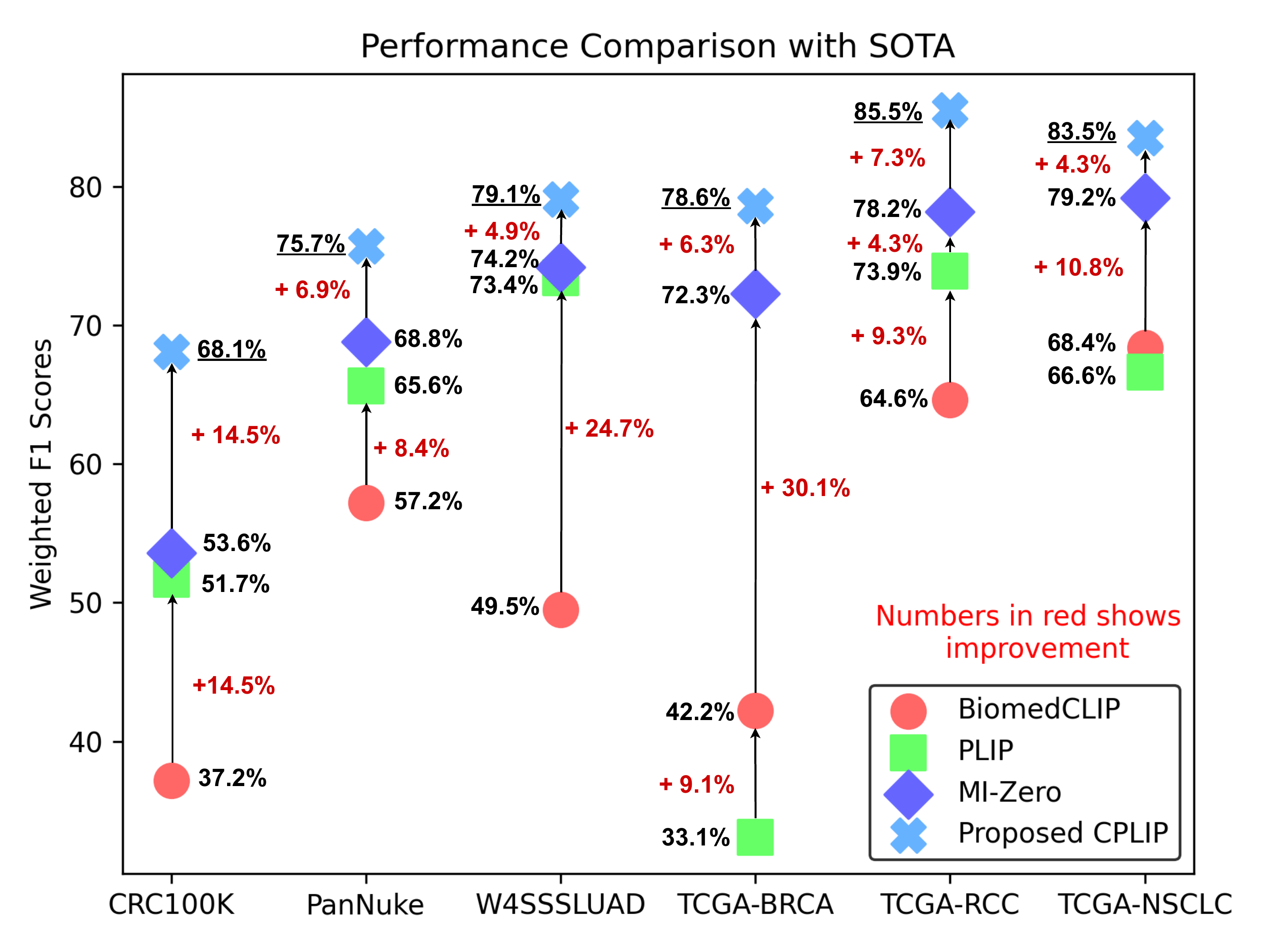}
\caption{Comparative analysis of zero-shot classification performance between the proposed CPLIP algorithm and existing SOTA methods such as BiomedCLIP \cite{zhang2023large}, PLIP \cite{huang2023visual}, and MI-Zero \cite{lu2023visual}. 
The weighted $F_{1}$ scores demonstrate CPLIP's substantial performance enhancements across six independent histology datasets. 
}
\label{fig_new}
\vspace{-.7cm}
\end{figure}

Translating these advances to computational pathology, VL models have transitioned from being novel to essential, enabling the fine-tuning of datasets considerably smaller than those typically used for VL pretraining \cite{huang2023visual, lu2023towards, lu2023harnessing, lai2023clipath}. Despite this progress, the scarcity of Whole Slide Images (WSIs) and diverse cancer morphologies poses a challenge for the zero-shot transfer capabilities of VL models, particularly for tasks like patch-based tissue recognition and WSI-level cancer subtyping, which are crucial during the inference phase \cite{lu2023towards}. Nevertheless, the successful deployment of VL models in classifying and analyzing WSIs underscores their significant role in revolutionizing the field of computational pathology.
%
\\
\indent
The use of VL models in classifying and analyzing WSIs has shown their impact on computational pathology \cite{lu2023towards}.
Lu  \textit{et al.} created a dataset of 33.48K histology image-caption pairs, which helped to fine-tune the CLIP model for cancer subtyping \cite{lu2023visual}. 
Huang \textit{et al.} collected about 208K histology images and texts from Medical Twitter to further fine-tune the CLIP model's ability for zero-shot classification and matching \cite{huang2023visual}.
Zhang \textit{et al.} also collected a heterogeneous dataset of 15 million image-text pairs, strengthening the CLIP model's training foundation \cite{zhang2023large}.
%
%
\begin{figure*}[t!]
\centering
\includegraphics[width=\linewidth]{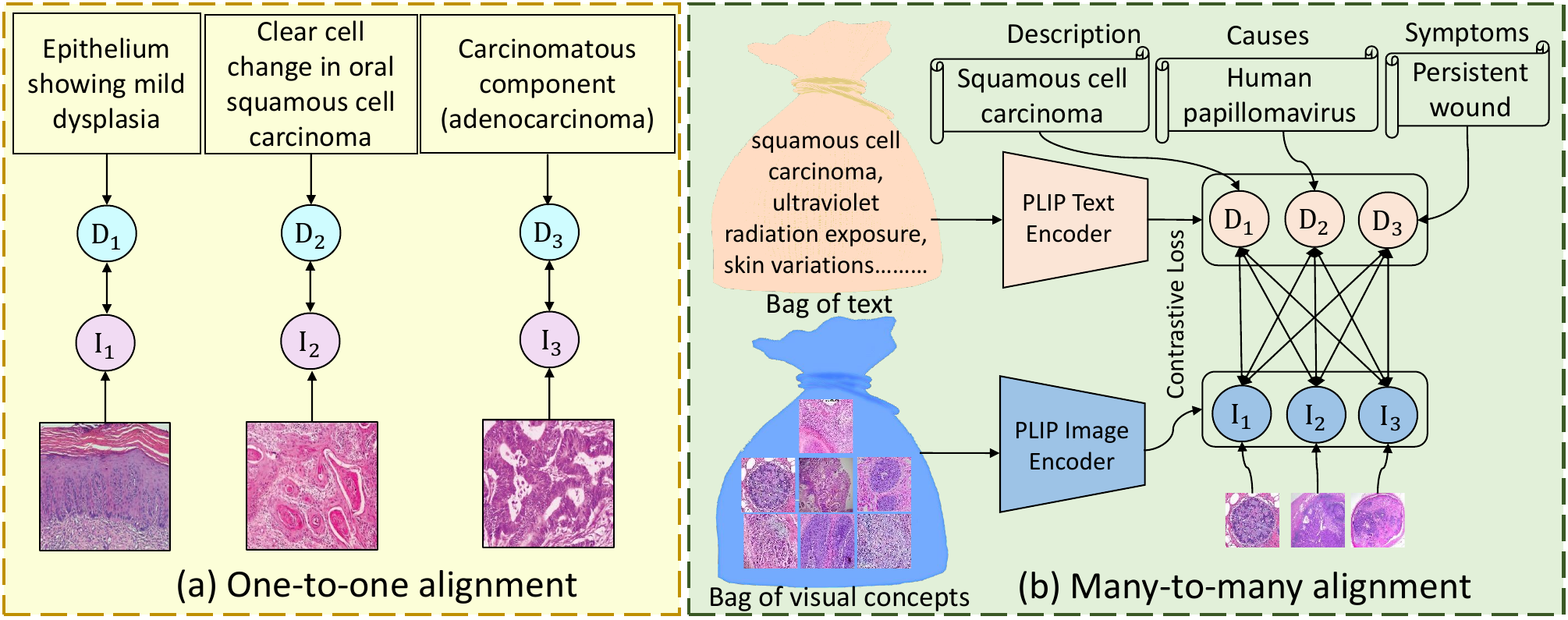}
\caption{
{\bf (a)} Displays the traditional one-to-one alignment in computational pathology VL models like PLIP \cite{huang2023visual}, BiomedCLIP \cite{zhang2023large}, and MI-Zero \cite{lu2023visual}, where each histology image is aligned with a single textual description during fine-tuning. {\bf (b)} Our proposed approach of many-to-many alignment, where bags of correlated texts are aligned with bags of correlated histology images during fine-tuning, offers a richer, interconnected data set for model training.}
\label{fig1}
\vspace{-.6cm}
\end{figure*}
\indent
\\
\indent
Textual prompts play a crucial role in improving VL models' performance. 
Yet, the tendency of these models to rely on just one phrase for each histology image might limit their zero-shot classification effectiveness \cite{huang2023visual, zhang2023large, lu2023visual}. They often use simple noun-based phrases, which may ignore detailed causes and symptoms of specific cancers. Introducing richer, more detailed prompts could provide VL models with a broader range of information during training, potentially improving their ability to classify and understand various cancer types.
To our knowledge, no existing histology VL models have incorporated such diverse textual prompts either during training or at the inference stage. 
Unlike existing methods focusing on aligning {\bf individual} textual and visual concepts, we propose the simultaneous alignment of numerous interrelated textual and visual concepts, as depicted in Figs. \ref{fig1} (a) \& (b).
%
\\
\indent
In this paper, we define ``comprehensiveness" as the incorporation of a broad array of textual descriptions for the same medical conditions, coupled with a diverse set of histology images for those conditions. This approach acknowledges that a single disease may be described differently by various medical professionals and can manifest in multiple ways across patients. Despite these variances, combining different descriptions and images provides a holistic view, enhancing the VL models' ability to make connections between symptoms, causes, and specific medical conditions.
%
\\
\indent
To generate ``comprehensive" textual prompts, we first compiled a pathology-specific dictionary cataloging various cancer types and related medical conditions, using a range of publicly available online glossaries. We then used an existing VL model \cite{lu2023visual} to select the most appropriate prompts for each histology image from this dictionary.
With GPT-3 \cite{brown2020language}, we transformed the selected prompts into five unique variations and identified three main causes and symptoms for each condition. 
Using the Pathology Language Image Pre-training (PLIP) model \cite{huang2023visual}, we matched these enhanced prompts with corresponding histology images from a Twitter dataset to enrich our visual database. 
The number of textual descriptions and images was capped at 17 and 21 to manage computational demands, though this limit can be adjusted according to resource availability.
%
\\
\indent
Using our extensive collection of textual prompts and visual content, we generated collections—or 'bags'—of textual descriptions and images through an unsupervised and automated process.
Images that match the prompts from our pathology dictionary are labeled as positive examples, while mismatches are negative. 
These collections are then used to fine-tune the CLIP model by adjusting the model's embeddings to align similar (positive) concepts and push away dissimilar (negative) ones. 
This method is aimed to enhance class-agnostic representations (refer to Fig. \ref{fig_new}). 
Our resulting fine-tuned model, called Comprehensive PLIP (CPLIP), is suited for various downstream zero-shot classification tasks.

\indent
This approach aligns with trends in AI that enhance interaction between language and visuals, much like VISPROG \cite{Gupta_2023_CVPR}, which translates language instructions into visual task actions. Similarly, our proposed CPLIP model integrates detailed textual and visual information to improve understanding in computational pathology.
%
In summary, our contributions include:
\begin{itemize}
\item 
Compilation of a dedicated dictionary for pathology-related prompts to facilitate the organized collection and application of comprehensive textual descriptions, improving model training and evaluation (Sec. \ref{sec:dictionary}).
%
\item
Development of comprehensive textual descriptions paired with multiple visual concepts to better align text and image embeddings (Secs. \ref{sec:bagoftext} \& \ref{sec:bagofimages}).
\item Advocacy for collective alignment of multiple textual descriptions and visual concepts (Sec. \ref{sec:loss}).
\item 
Demonstrated superior zero-shot performance by our model on different datasets, highlighting the benefits of the \enquote{comprehensiveness} approach to boosting VL models for classification and segmentation in computational pathology (Sec. \ref{sec:results}).
%
\end{itemize}

\section{Related Work}
\label{sec:relatedwork}
In the effort to advance computational pathology through various tasks like histology image classification, segmentation, and survival prediction, numerous methods have been proposed \cite{srinidhi2021deep,chen2022semi}.
These methods can be broadly categorised as weakly-supervised \cite{ilse2018attention, shao2021transmil}, self-supervised \cite{wang2022transformer, koohbanani2021self}, and Vision-Language (VL) supervised \cite{huang2023visual, lu2023visual, zhang2023large}.
\\
\noindent \textbf{(i) Weakly-supervised Learning Methods (WSL)} use data with labels at a broad level, without needing detailed annotations for every instance. 
In computational pathology, Multiple Instance Learning (MIL) has evolved as a popular paradigm for WSI classification.
Examples include ABMIL \cite{ilse2018attention}, TransMIL \cite{shao2021transmil}, DSMIL \cite{shao2021transmil}, CLAM \cite{lu2021data}, and DTFD-MIL \cite{zhang2022dtfd}. 
%
%
{\it In contrast to this paradigm, our VL-based algorithm does not require any label during the training rather it uses pathology-specific language supervision.}

%

\noindent 
\textbf{(ii) Self-supervised Learning Methods} in computational pathology learn from the data itself without using labels, using pretext tasks to boost downstream task performance.
Key in this area is contrastive learning-based methods, which focus on distinguishing between similar and contrasting instances within the data. 
By using contrastive loss, these methods train models to discern augmentation-invariant features crucial for tasks like classification and anomaly detection. Examples include CTransPath for histology image classification \cite{wang2022transformer}, H2T \cite{vu2023handcrafted}, HIPT \cite{chen2022scaling}, and \cite{koohbanani2021self}. These techniques help models capture essential inherent data characteristics, enhancing performance on various computational pathology applications.
{\it Our approach goes beyond contrastive learning methods by not only aligning bags of images but also aligning bags of texts and additional strategies to enhance model performance.}
%
%
%
\\
\noindent 
\textbf{(iii) Learning with Pathology Language Supervision Methods} integrate textual descriptions with visual data to pre-train deep models.
Adhering to the conventional VL model training approach, these methodologies leverage paired visual-textual data within a contrastive learning framework to ensure that representations of similar visual-textual concepts are drawn closer together, while divergent ones are distanced \cite{lu2023towards, chen2022semi, radford2021learning}.
Recently, the VL paradigm has been extended to zero-shot classification and segmentation tasks, introducing models like PLIP \cite{huang2023visual}, CONCH \cite{lu2023towards}, MI-Zero \cite{lu2023visual}, and BiomedCLIP \cite{zhang2023large}.
These innovations have brought forth new datasets containing descriptions of histology images and languages to pre-train architectures resembling CLIP \cite{radford2021learning}. 
A limitation of these models is their potential inability to generalize well across different datasets due to the training dataset-specific biases consisting of paired textual-visual concepts.
%
Also most of these approaches primarily focus on aligning single textual and visual representations.
In contrast, we propose the engagement of comprehensive visual and textual data to concurrently align multiple correlated positive visual-textual concepts. 
We argue that such an expansive and robust alignment significantly elevates performance across a spectrum of computational pathology tasks.

\section{Proposed Methodology}
\label{sec:method} 
\indent
In this work, we propose the Comprehensive Pathology Language Image Pre-training (CPLIP) algorithm. 
This algorithm effectively uses a collection of unlabeled histology images, paired with a predefined comprehensive pathology prompt dictionary, to fine-tune the CLIP model without any ground truth annotations (neither at the image level nor at the text level).
The purpose is to tailor CLIP to a diverse range of histology data gathered from various sources. This enhances its ability for zero-shot transfer across different computational pathology tasks, especially for unfamiliar tissue categories not encountered during the training phase.
We represent the comprehensive pathology prompts dictionary as \textit{V} and denote the collection of unlabeled histology images as $\textit{H} = {\{\textit{h}_{j}\}_{j=1}^{n_h} }$, where $n_{h}$ indicates the total count of these histology images.
\\
\indent
Fig. \ref{fig2} illustrates the process of constructing the bag of textual descriptions and the bag of visual concepts within our CPLIP framework. It depicts the primpary phases, including the construction of a predefined pathology prompt dictionary and the aggregation of corresponding textual descriptions, followed by the formation of visual concepts. These elements are integral to our many-to-many contrastive learning approach, which seeks to align positive visual-textual pairs and separate negative ones. The details of these processes are discussed in the following sections.
%
%
\begin{figure*}[t!]
\centering
\includegraphics[width=\linewidth]{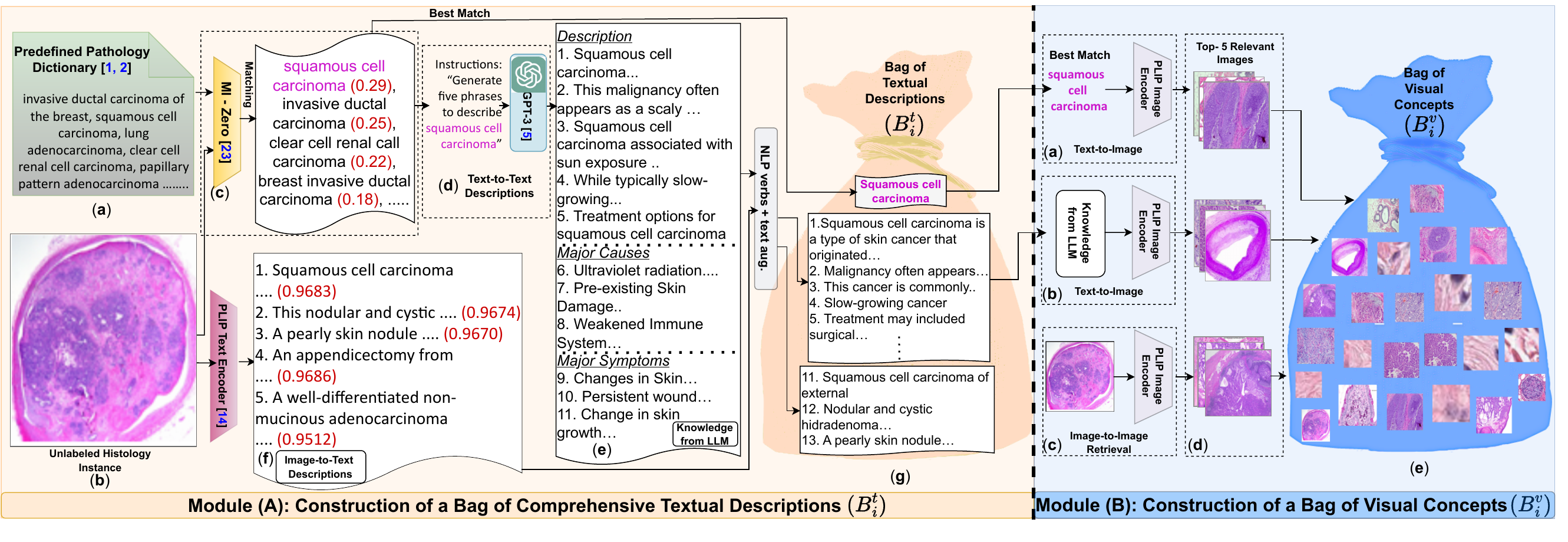}
\caption{Diagram outlining the construction of comprehensive textual descriptions and visual concept bags. 
(A) illustrates the construction process of the textual description bag, while (B) shows the procedure for constructing the visual concept bag. 
Within (A), there are three primary steps: using MI-Zero to identify the best text match, leveraging GPT-3 to enrich the textual descriptions of the best-matched text, and employing the PLIP text encoder to generate more in-depth descriptions of the input unlabeled histology image.
Within (B), there are also three primary steps: (a) using PLIP to identify the best-matching images, (b) leveraging PLIP to enrich the histology images of the best-matched textual descriptions, and (c) employing the PLIP to retrieve relevant histology images of the input unlabeled histology image.
}
\label{fig2}
\end{figure*}
\subsection{Predefined Pathology Dictionary (Fig. \ref{fig2} A(a))}
\label{sec:dictionary}
The ARCH dataset is the only publicly accessible histology image-caption pairing \cite{gamper2019pannuke}.
This dataset has been used in MI-Zero \cite{lu2023visual}.
This method, however, restricts them to paired image-text data, which might not be comprehensively available to the public.
To address this limitation, we propose a set pathology prompt dictionary.
This serves as a foundational prompt to extract more comprehensive images and textual descriptions in subsequent phases.
\\
\indent
We have created a strong dictionary tailored to histology descriptions, which includes terms commonly used by expert pathologists to describe various cancer forms, related medical conditions, and their prognoses through histology images. 
To generate this resource, we merged cancer glossaries from esteemed institutes \cite{cancer1, cancer2} manually refining the collected data to form a more precise pathology-specific dictionary. 
Our experiments compare the effectiveness of these two vocabularies, assessing the outcomes of each. 
The combined dictionary holds 500 varied prompts, incorporating 1,500 terms covering the range of cancer types and morphologies for diagnosis. 
After refinement (cleaning and filtering), the dictionary has 200 wide-ranging and in-depth prompts totaling 700 terms. 
This refinement process first removes irrelevant prompts, like those not directly connected to a histology image. 
It then omits non-histopathology prompts, sidestepping those related to radiology, X-rays, CTs, and so on. 
Every prompt is thereafter denoted with a suitable acronym and corresponding description.
The refined predefined prompts dictionary is designed to cover major cancer types and morphologies across various tissue types. 
We have provided it as supplementary material in this paper and intend to release it to the public.

\subsection{Building a Comprehensive Textual Descriptions Bag (Fig. \ref{fig2} A(g))}
\label{sec:bagoftext}
Given the collection of input unlabelled histology images, denoted as $\textit{H}$, and the predefined pathology prompts dictionary, $\textit{V}$, we generate a detailed textual description bag, $B_{i}^{t}$, for each image $h_j \in \textit{H}$. 
This process uses three distinct textual sources: MI-Zero \cite{lu2023visual}, GPT-3 \cite{brown2020language}, and PLIP \cite{huang2023visual}. 
Each source is elaborated on below.
\vspace{-.5cm}
\subsubsection{Matching with MI-Zero (Fig. \ref{fig2} A(c))}
\label{sec::matching}
While the widely used visual text encoder CLIP is trained on generic data, our work necessitates a domain-specific VL encoder. 
With limited options available, we opted for the MI-Zero model \cite{lu2023visual}. 
This model, recently launched, is trained on matched histopathological image-caption data. 
MI-Zero includes a visual encoder, represented as $f (\cdot;\theta)$, and a text encoder, $g (\cdot;\phi)$, both of which compute image and text embeddings, respectively. 
For a given histology image $h_{j}$, we identify its most related prompts from the predefined dictionary \textit{V}  using the formula:
\begin{equation}
\hat{v}_{i}=\argmax_{v_{i} \in V} \textrm{sim} (f(h_{j}), g(v_{i})), 
\label{eqn1}
\vspace{-.08cm}
\end{equation}
\noindent Here, $\textrm{sim}(x,t)=x^\top t/(||x||~||t||)$ denotes the cosine similarity measure. 
The resulting $\hat{v}_{i}$ is then added to the textual descriptions bag, $B_{i}^{t}$.
Fig. \ref{fig2} (A) provides a visual representation of this process. 
It begins with the predefined pathology prompt dictionary (shown in Fig. \ref{fig2} A(a)) and an unlabelled histology image (Fig. \ref{fig2} A(b)). 
From here, we identify the top five matching prompts (Fig. \ref{fig2} A(c)).
Of these, only the best-matching text, termed \enquote{squamous cell carcinoma}, is chosen and added to the bag $B_{i}^{t}$ (See Fig. \ref{fig2} A(g)). 
For more examples of closely matched prompts, refer to our supplementary material.
%
\subsubsection{GPT-3 for Comprehensive Textual Descriptions (Fig. \ref{fig2} A(d))}
To derive multiple descriptions of the top-ranked textual prompt from the previous process (Sec. \ref{sec::matching}), we can turn to Large Language Models (LLM) like GPT-3 \cite{brown2020language}. 
Such models have demonstrated strong capabilities in various linguistic tasks \cite{wang2023survey}. 
By inputting the highest ranked prompt $\hat{v}_{i}$ into LLM, we ask it to produce five alternate descriptions based on its extensive linguistic knowledge. 
Additionally, we generate three primary etiologies/causes and three dominant symptoms related to the top-ranked prompt using the LLM. 
An example presented in Fig. \ref{fig2} A(e) reveals five unique descriptions for the top-rated prompt \enquote{squamous cell carcinoma}.
Some descriptions include \enquote{squamous cell carcinoma is a common form of skin cancer} and \enquote{squamous cell carcinoma malignancy often appears as a scaly, red patch...} and so on. 
The model also provides potential causes like \enquote{prolonged exposure to ultraviolet radiation} and noticeable symptoms like \enquote{skin alterations} and \enquote{lingering wound}.
Together with the identified pathology prompt \enquote{squamous cell carcinoma}, this brings forth twelve varied textual descriptions that will be considered during the formulation of our bag $B_{i}^{t}$.
Moreover, our text augmentation includes the lemmas of verbs, helping the model to treat different verb forms as the same action.
Additional illustrations are made available in the supplementary material.
%
%
\subsubsection{Image-to-Text Description with PLIP (Fig.\ref{fig2} A(f))}
In this phase, we use the PLIP model to identify relevant text descriptions linked to given unlabeled histology images, pulling information from the vast Medical Twitter dataset \cite{huang2023visual}.
From this, we select the top five most appropriate descriptions and add them to our textual bag, $B_{i}^{t}$.
The PLIP model consists of both visual and text encoders, specifically adapted based on the large-scale medical Twitter dataset. 
With PLIP's assistance, we can integrate descriptions from a variety of sources into our textual bag, $B_{i}^{t}$. 
Fig. \ref{fig2} A(f) displays the top five descriptions matched to the unlabeled histology image shown in Fig. \ref{fig2} A(a).

\begin{table*}[t!]
\caption{Ablations 1-3: Zero-shot classification performance comparison in terms of weighted $F_{1}$ score using different heterogeneous textual descriptions. 95\% Confidence Interval (CI) is included in parentheses.}
\begin{center}
\makebox[\linewidth]{
\scalebox{0.95}{
\begin{tabu}{|c|c|c|c|c|c|}
\tabucline[1.5pt]{-}
Ablation Study&D500+GPT-3+PLIP&D200+GPT-3+PLIP&D200 only&D200+GPT-3&GT+GPT-3+PLIP\\\tabucline[1.5pt]{-}
CRC100K&0.804(0.791,0.815)&
\underline{0.844}(
\underline{0.833},
\underline{0.856})&
0.697(0.682,0.704)&0.774(0.752,0.703)&\textbf{0.861(0.852,0.874)}\\\tabucline[0.5pt]{-}
DigestPath&0.842(0.833, 0.859)&
\underline{0.903}(
\underline{0.891}, 
\underline{0.915})&0.734(0.707,0.764)&0.831(0.820,0.834)&\textbf{0.912 (0.904,0.922)}\\\tabucline[0.5pt]{-}
SICAP&0.441(0.401,0.485)&
\underline{0.511}(
\underline{0.498},
\underline{0.526})&0.292(0.276,0.317)&0.422(0.375,0.475)&\textbf{0.533(0.508,0.571)}\\\tabucline[0.5pt]{-}
W4SSSLUAD&0.801(0.772,0.835)&
\underline{0.882}(
\underline{0.876},
\underline{0.894})&0.644(0.605,0.683)&0.716(0.685,0.743)&\textbf{0.891(0.876,0.916)}\\\tabucline[0.5pt]{-}
PanNuke&0.761(0.744,0.786)&
\underline{0.811}(
\underline{0.799},
\underline{0.827})&0.685(0.613,0.749)&0.761(0.753,0.772)&\textbf{0.841(0.815,0.873)}
\\\tabucline[1.5pt]{-}
\end{tabu}
}}
\end{center}
\label{ablation1}
\end{table*}

\subsection{Compiling Visual Concepts Bag (Fig. \ref{fig2} B(e))}
\label{sec:bagofimages}
The visual concepts repository, denoted as $B_{i}^{v}$, is constructed based on the comprehensive textual descriptions sourced from the textual bag $B_{i}^{t}$ and the unlabeled histology image $h_{j}$.
This process consists of two primary stages, as depicted in Fig. \ref{fig2} (B).
\subsubsection{Image Retrieval with PLIP Based on Textual Prompts (Fig. \ref{fig2} B(a)-(b))}
Starting with a top-matched prompt from MI-Zero matching, we identify several histology images that match this prompt using PLIP's image and text encoders. 
Similarly, using a set of textual descriptions from LLM, we find related histology images that go with each description through the PLIP model. 
It is important to note that our approach uses only the pre-trained PLIP model without any extra fine-tuning.
For example, with the prompt \enquote{squamous cell carcinoma} as our top match, and using textual information from LLM (as shown in Fig. \ref{fig2} A(e)), we were able to identify a total of 16 images that were relevant.
\subsubsection{PLIP Image-to-Image Retrieval (Fig. \ref{fig2} B(c))}
With the unlabelled histology image, $h_{j}$, we retrieve the five most related images using PLIP's image encoder from the Medical Twitter dataset. 
When these are added to $B_{i}^{v}$, the total comes to 21 visual concepts. 
The top five images, as an example (Fig. \ref{fig2} B(c)), can be viewed in the supplementary material.
\subsection{Textual and Visual Bags Pruning}
\label{sec:prunning}
Considering the textual descriptions in $B_{i}^{t}$ come from various sources, there is a chance some may not be as relevant.
To improve the $B_{i}^{t}$ bag quality, we make sure each description $t_{i,n} \in B_{i}^{t}$ closely matches with input image $h_{j}$, exceeding a specific similarity value $\textrm{sim} (f(h_{j}), g(v_{i}))\ge \delta_{t}$.
Adjusting this $\delta_{t}$ value can either reduce the number of descriptions in $B_{i}^{t}$ (if the value is higher) or keep most of them (if it is lower).
Since $B_{i}^{v}$ is constructed using the pruned textual bag and the PLIP model, the pruning applied on the $B_{i}^{t}$ consequently reflects in the $B_{i}^{v}$. 
Please note no further pruning is applied on bag $B_{i}^{v}$.
\subsection{MIL-based Contrastive Loss (Fig. \ref{fig1} (b))}
\label{sec:loss}
To fine-tune the PLIP model, we use the Multiple Instance Learning-Noise Contrastive Estimation (MIL-NCE) loss introduced in \cite{miech2020end}. Contrary to the original MIL-NCE design that aligns a single positive text with a single positive video, our algortihm CPLIP connects a bag of text, $B_{i}^{t}$, with a corresponding set of visual bags, $B_{i}^{v}$ (the specific sequence of items from the bags is inconsequential).
This approach facilitates the association of multiple textual descriptions with multiple histology images. 
Our defined MIL-NCE loss function is presented as follows:
\begin{equation}
      \mathcal{L}=  -\frac{1}{B} \sum_{i} log\Bigg[\frac{  \sum_{m} \sum_{n}  \exp{\Big( f (v_{i,m})^\top g (t_{i,n})/\sigma \Big)}}{ \sum_{m} \sum_{j} \sum_{n}   \exp{ \Big(f (v_{i,m})^\top g (t_{j,n})/\sigma\Big)}} \Bigg],
        \label{eqn2}
\end{equation}

\noindent Here, $v_{i,m} \in B_{i}^{v}$, $t_{j,n} \in B_{j}^{t}$, $0<n \le n_{bag}$, and $0<m \le m_{bag}$. $m_{bag}$ represents the size of $B_{i}^{v}$, while $n_{bag}$ denotes the size of $B_{i}^{t}$. $B$ and $\sigma$ indicate the batch size and the constant temperature parameter, respectively.
\subsection{Zero-shot Transfer for Histology Landscape}
Radford \textit{et al.} have introduced a method that uses prompts for zero-shot classification \cite{radford2021learning}.
In this method, class names are converted into prompts by attaching them to specific keyword templates. 
For instance, the class name \enquote{Tumor Adenocarcinoma} is expanded using the template \enquote{An H \& E image of \{\}}.
Subsequently, the trained text encoder calculates the embeddings of these prompts.
Meanwhile, the trained visual encoder deduces the embeddings of test images. 
These embeddings are normalized using $\ell_{2}$, and their similarity is measured using the cosine similarity measure. 
The labels of the test images are determined based on the highest similarity scores.
Given the variance in the performance of different prompts, we expand the prompt generation process. 
We use a set of templates tailored for pathology and introduce alternative names for each class, drawing inspiration from earlier studies \cite{lu2023visual, lu2023towards}. 
When making inferences, the various prompts for each class are combined by averaging their embeddings. 
Our experiments present results both with and without the merging of prompts.

\section{Experiments}
\label{sec:results}
We conduct several experiments to evaluate the proposed CPLIP algorithm, including tile-level zero-shot classification, WSI-level zero-shot classification, and zero-shot segmentation of gigapixel WISs. 
For the tile-level zero-shot classification, we use five independent datasets: CRC100K \cite{kather2019predicting}, WSSS4LUAD \cite{han2022wsss4luad}, PanNuke \cite{gamper2019pannuke}, DigestPath \cite{da2022digestpath}, and SCIAP \cite{silva2021self}.
For the WSI-level zero-shot classification, we use four datasets: CAMELYON-16 (CAM16) \cite{bejnordi2017diagnostic} and others from The Cancer Genome Atlas (TCGA) including BRCA, RCC, and NSCLC \cite{tomczak2015review}. 
Finally, for the zero-shot segmentation, we use the SICAP and DigestPath datasets. 
Through these diverse experiments spanning tiles, WSIs, and segmentation tasks, we comprehensively assess the performance of the proposed CPLIP method.

\begin{table}[t!]
\caption{Ablation 5: Zero-shot classification performance comparison in terms of weighted $F_{1}$ score for a bag of text vs. a bag of visual concepts. 95\% CI is included in parentheses.}
\begin{center}
\makebox[\linewidth]{
\scalebox{0.70}{
\begin{tabu}{|c|c|c|c|}
\tabucline[1.5pt]{-}
Ablation Study&Text bag ($B^{t}$)&Visual bag ($B^{v}$)&Proposed\\\tabucline[1.5pt]{-}
CRC100K&\underline{0.761}(\underline{0.753},\underline{0.774})&0.744(0.723,0.765)&\textbf{0.844(0.833,0.856)}\\\tabucline[0.5pt]{-}
DigestPath&0.854(0.831, 0.872)&\underline{0.861}(\underline{0.852},\underline{0.871})&\textbf{0.903(0.891,0.915)}\\\tabucline[0.5pt]{-}
SICAP&\underline{0.477}(\underline{0.465},\underline{0.487})&0.471(0.451,0.495)&\textbf{0.511(0.498,0.526)}\\\tabucline[0.5pt]{-}
W4SSSLUAD&0.772(0.752,0.793)&\underline{0.786}(\underline{0.772},\underline{0.796})&\textbf{0.882(0.876,0.894)}\\\tabucline[0.5pt]{-}
PanNuke&\underline{0.766}(\underline{0.734},\underline{0.795})&0.756(0.723,0.785)&\textbf{0.811(0.799,0.827)}\\\tabucline[1.5pt]{-}
\end{tabu}
}}
\end{center}
\label{ablation5}
\end{table}

\begin{table*}[t!]
\caption{Ablation 4: Zero-shot classification performance in terms of bags pruning using weighted $F_{1}$ score with 95\% CI.}
\begin{center}
\makebox[\linewidth]{
\scalebox{0.80}{
\begin{tabu}{|c|c|c|c|c|c|c|}
\tabucline[1.5pt]{-}
Matching&ratio $\delta_{t}$&CRC100K&DigestPath&SICAP&W4SSSLUAD&PanNuke\\\tabucline[1.5pt]{-}
MI-Zero matching&100\%&0.806(0.791,0.813)&0.871(0.867,0.884)&0.446(0.402,0.485)&0.871(0.864,0.885)&0.798(0.775,0.817)\\\tabucline[0.5pt]{-}
MI-Zero matching&90\%&
\textbf{0.844(0.833,0.856)}&
\textbf{0.903(0.891,0.915)}&0.488(0.474,0.493)&
\textbf{0.882(0.876,0.894})&
\textbf{0.811(0.799,0.827)}\\\tabucline[0.5pt]{-}
MI-Zero matching&70\%&
\underline{0.833}(
\underline{0.821},
\underline{0.841})&
\underline{0.896}(
\underline{0.861},
\underline{0.928})&
\textbf{0.511(0.498,0.526)}&
\underline{0.880}(
\underline{0.861},
\underline{0.890})&0.804(0.791,0.813)\\\tabucline[0.5pt]{-}
MI-Zero matching&50\%&0.829(0.814,0.838)&0.883(0.864,0.905)&
\underline{0.507}(
\underline{0.472},
\underline{0.534})&0.875(0.866,0.886)&
\underline{0.805}(
\underline{0.775},
\underline{0.836})\\\tabucline[0.5pt]{-}
MI-Zero matching&30\%&0.827(0.831,0.858)&0.881(0.876,0.898)&0.501(0.485,0.525)&0.873(0.854,0.895)&0.803(0.786,0.825)\\\tabucline[1.5pt]{-}
\end{tabu}
}}
\end{center}
\label{ablation3}
\end{table*}

\subsection{Training and Implementation Details}
In histopathology, the ARCH dataset \cite{gamper2021multiple} is the only widely available image-text paired dataset, containing 8,617 pairs from clinical and research pathology articles. 
We fine-tuned our CPLIP algorithm on this dataset, extending it to around 180,000 images and 146,000 textual descriptions without using their paired texts. 
This unpaired many-to-many image-text alignment is a novelty compared to the paired data approach used by MI-Zero \cite{lu2023visual}. 
Our fine-tuning process involved various architectures, leveraging domain-specific and general models, with modifications to suit our many-to-many alignment needs. We used a batch size of 256 for 50 epochs, applying specific filtering thresholds to refine the data further. While single prompts were used for reporting results, additional details on the use of merged prompts and further implementation details are provided in the supplementary material section.
\subsection{Datasets and Evaluation Metrics}
We used nine independent publicly available {\bf computational pathology datasets} for classification and segmentation tasks (more detailed descriptions of each dataset are provided in the supplementary material), spanning diverse cancer types and image modalities including \textbf{(i)} CRC100K \cite{kather2019predicting} colorectal cancer dataset used for zero-shot tile classification on 7,180 test images across nine tissue types; \textbf{(ii)} WSSS4LUAD \cite{han2022wsss4luad} lung adenocarcinoma dataset used for zero-shot tumor vs. normal classification on 3,028 test images; \textbf{(iii)} SICAP \cite{silva2021self} prostate cancer dataset used for zero-shot classification on 2,122 test images with 4 Gleason pattern labels; \textbf{(iv)} PanNuke \cite{gamper2019pannuke} diverse tissue dataset used for zero-shot tumor vs. normal classification on 1,888 test images with 19 tissue types; \textbf{(v)} DigestPath \cite{da2022digestpath} colonoscopy tissue dataset used for zero-shot tumor vs. normal tile classification on 18,814 test images; \textbf{(vi)} Camelyon 16 (CAM16) \cite{bejnordi2017diagnostic} breast cancer dataset used for zero-shot slide classification on 130 test slides; and \textbf{(vii-ix)} TCGA \cite{tomczak2015review} invasive BRCA, RCC, and NSCLC datasets used for zero-shot slide classification on 75 slides per class. 
In summary, these diverse ranges of computational pathology datasets are used to evaluate zero-shot classification and segmentation performance across testing sets ranging from thousands of image tiles to hundreds of WSIs.
Our {\bf evaluation metrics} include balanced accuracy, weighted $F_{1}$ score, and AUCROC for classification tasks, and the Dice score, precision, and recall for segmentation tasks, in line with current SOTA VL methods \cite{lu2023towards, lu2023visual, huang2023visual}. Balanced accuracy is calculated by averaging the recall of each class.
%
\subsection{SOTA Methods for Comparison}
We compared the performance of our proposed CPLIP algorithm with several recently proposed SOTA methods on zero-shot classification and segmentation tasks for histopathology images. 
We included five recently proposed VL-based methods in our comparison: baseline CLIP \cite{radford2021learning}, PLIP \cite{huang2023visual}, MI-Zero \cite{lu2023visual}, BiomedCLIP \cite{zhang2023large}, and CONCH \cite{lu2023towards}.
To ensure a fair comparison, we used the official source code for all methods and kept the same settings for testing splits and inference prompts, except for CONCH, whose source code is not yet available.

\subsection{Ablation Studies}
All ablation studies use CTransPath \cite{wang2022transformer} as the image encoder and BioClinicalBert \cite{alsentzer2019publicly} to initialize the text encoder, with performance reported using merged prompts.
For more details and ablation studies, see supplementary material.

\noindent \textbf{1. Cleaned vs. Uncleaned Pathology Prompts dictionary.}
This experiment compares the performance of zero-shot classification using the original unsupervised pathology prompts dictionary consisting of 500 prompts (D500) vs. a manually cleaned pathology prompts dictionary containing 200 prompts (D200) (see Sec. \ref{sec:dictionary}). 
As shown in Table \ref{ablation1}, D200$+$GPT-3$+$PLIP achieved better performance on five datasets compared to D500$+$GPT-3$+$PLIP.
\textit{This indicates that a smaller, curated dictionary of 200 cleaned pathology prompts yields better zero-shot classification results than a larger, uncleaned noisy set of 500 prompts.}

\noindent \textbf{2. Effect of paired image-text supervision.}
This experiment removed the pathology prompts dictionary step and used the ARCH paired text as the best match prompt in Sec. \ref{sec:dictionary}.
The paired text data was then used to construct the textual bag using GPT-3 and PLIP text encoder to obtain a similar textual bag as in Sec. \ref{sec:bagofimages}.
The zero-shot classification performance of this strategy (GT+GPT-3+PLIP) is also shown in Table \ref{ablation1}.
The results showed that ground truth text-based results were better than the unsupervised dictionary results.
\textit{This indicates that using ARCH's paired text data to construct textual bags via GPT-3 and PLIP text encoder, as opposed to an unsupervised dictionary, improves zero-shot classification performance.}

\noindent \textbf{3. Importance of heterogeneous textual and visual resources.}
We conducted experiments using only D200 pathology dictionary (using a single best match prompt and a single image), D200+GPT-3 (using 12 textual descriptions and 12 images), and D200+GPT-3+PLIP (using 17 textual descriptions and 21 images). 
\textit{The results in Table \ref{ablation1} show that adding more textual resources during training improves performance on all datasets.}

\noindent \textbf{4. Effect of Bags Pruning.}
In this experiment, the textual bags ($B^{t}$) were pruned to retain 90$\%$, 70$\%$, 50$\%$, and 30$\%$ of the best matching textual descriptions with the input image using cosine similarity (see Sec. \ref{sec:prunning}).
The corresponding visual bags ($B^{v}$) were also pruned subsequently. 
A 100$\%$ bag means no pruning, and it may contain some noisy text. 
As shown in Table \ref{ablation3}, the best zero-shot classification performance over four datasets was observed for $\delta_{t}=90\%$.
\textit{Further pruning reduced performance due to data reduction, which resulted in reduced heterogeneity.}

\noindent \textbf{5. Which Bag is more important?} 
We conducted two experiments to compare the importance of the textual and visual bags for contrastive training. 
In the first experiment, we used a bag of text along with the input image ($B_{j}^{t} + h_{j}$). 
In the second experiment, we used only the bag of visual concepts $B^{v}$.
Both experiments observed performance degradation compared to the proposed $B^{t}+ B^{v}$ based training, as shown in Table \ref{ablation5}.
\textit{This suggests that both the textual and visual bags are important for achieving good performance.}

\begin{table*}[t!]
\caption{Tile-level zero-shot classification performance comparison using single prompt in terms of balanced accuracy, weighted $F_{1}$, and AUROC scores with other SOTA methods across five datasets. 
The CPLIP algorithm outperforms existing models. For the WSSS4LUAD dataset, CONCH used a different split, denoted by an asterisk ($^{*}$).}
\begin{center}
\makebox[\linewidth]{
\scalebox{0.80}{
\begin{tabu}{|c|c|c|c|c|c|}
\tabucline[1.5pt]{-}
Methods&CRC100K&DigestPath&SICAP&WSSS4LUAD&PanNuke\\\tabucline[1.5pt]{-}
CLIP baseline \cite{radford2021learning}&0.234$|$0.185$|$0.727&0.11$|$0.030$|$0.203&0.231$|$0.139$|$0.201&0.451$|$0.481$|$0.705&0.322$|$0.352$|$0.683\\\tabucline[0.5pt]{-}
BiomedCLIP \cite{zhang2023large}&0.422$|$0.372$|$0.859&0.591$|$0.622$|$0.781&0.381$|$0.361$|$0.506&0.466$|$0.495$|$0.698&0.522$|$0.572$|$0.711\\\tabucline[0.5pt]{-}
PLIP \cite{huang2023visual}&0.520$|$0.517$|$0.879&0.815$|$0.832$|$0.901&0.319$|$0.255$|$0.603&0.702$|$0.734$|$0.822&0.629$|$0.656$|$\underline{0.805}\\\tabucline[0.5pt]{-}
MI-Zero \cite{lu2023visual}&0.544$|$0.536$|$0.872&\underline{0.822}$|$\underline{0.811}$|$\underline{0.911}&0.308$|$\underline{0.251}$|$\underline{0.605}&\underline{0.722}$|$\underline{0.742}$|$\underline{0.805}&\underline{0.659}$|$\underline{0.688}$|$0.755\\\tabucline[0.5pt]{-}
CONCH \cite{lu2023towards}&\underline{0.566}$|$\underline{0.542}$|$\underline{0.901}&-&\underline{0.349}$|$0.245$|$-&0.598$^{*}$$|$0.590$^{*}$$|$0.795$^{*}$&-\\\tabucline[0.5pt]{-}
Proposed CPLIP &\textbf{{0.701}}$|$\textbf{{0.681}}$|$\textbf{{0.922}}&\textbf{{0.835}}$|$\textbf{{0.856}}$|$\textbf{{0.933}}&\textbf{0.366}$|$\textbf{{0.388}}$|$\textbf{{0.711}}&\textbf{{0.778}}$|$\textbf{{0.791}}$|$\textbf{{0.836}}&\textbf{{0.681}}$|$\textbf{{0.757}}$|$\textbf{{0.835}}\\\tabucline[1.5pt]{-}
\end{tabu}
}}
\end{center}
\label{table8}
\end{table*}

\begin{table*}[t!]
\caption{WSI-level zero-shot classification performance comparison  using single prompts, in terms of balanced accuracy, weighted $F_{1}$, and AUROC on four datasets. 
Both our Out-of-Domain (OoD) CPLIP$_1$ and In-Domain (InD) CPLIP$_2$ outperform across all metrics.}
\begin{center}
\makebox[\linewidth]{
\scalebox{0.70}{
\begin{tabu}{|c|c|c|c|c|c|c|}
\tabucline[1.5pt]{-}
Models (Single prompts)&Image encoder pretraining&Text encoder pretraining&CAM16&TCGA-BRCA&TCGA-RCC&TCGA-NSCLC\\\tabucline[1.5pt]{-}
CLIP baseline \cite{radford2021learning}&ViT-B/16-224&GPT-2/77&0.134$|$0.175$|$0.325&0.512$|$0.328$|$0.551&0.321$|$0.178$|$0.578&0.496$|$0.358$|$0.536\\\tabucline[0.5pt]{-}
BiomedCLIP \cite{zhang2023large}&ViT-B/16-224&PMB/256&0.311$|$0.377$|$0.545&0.527$|$0.422$|$0.761&0.677$|$0.646$|$0.872&0.699$|$0.684$|$0.851\\\tabucline[0.5pt]{-}
PLIP \cite{huang2023visual}&ViT-B/32-224&GPT/347&0.399$|$0.416$|$0.681&0.451$|$0.331$|$0.611&0.726$|$0.739$|$\underline{0.915}&0.676$|$0.666$|$0.781\\\tabucline[0.5pt]{-}
MI-Zero \cite{lu2023visual}&CTransPath/224&BioClinicalBert/512&0.456$|$0.461$|$\underline{0.755}&\underline{0.781}$|$\underline{0.723}$|$0.856&\underline{0.805}$|$0.782$|$0.881&0.802$|$0.792$|$0.866\\\tabucline[0.5pt]{-}
CONCH \cite{lu2023towards}&ViT-B/16-256&HistPathGPT/512&-&0.643$|$0.600$|$\underline{0.873}&0.796$|$\underline{0.797}$|$\textbf{0.961}&\underline{0.807}$|$\underline{0.803}$|$\underline{0.915}\\\tabucline[0.5pt]{-}
CPLIP$_{1}$ (Ours)&ViT-B/16-224 (OoD)&GPT-2/77 (OoD)&\underline{0.502}$|$\underline{0.477}$|$0.705&0.500$|$0.544$|$0.722&0.754$|$0.749$|$0.865&0.761$|$0.788$|$0.821\\\tabucline[0.5pt]{-}
CPLIP$_{2}$ (Ours)&PLIP-ViT-B/32-224 (InD)&PLIP-GPT/347 (InD)&\textbf{0.591}$|$\textbf{0.587}$|$\textbf{0.827}&\textbf{0.824}$|$\textbf{0.786}$|$\textbf{0.889}&\textbf{0.844}$|$\textbf{0.855}$|$0.926&\textbf{0.854}$|$\textbf{0.835}$|$\textbf{0.936}\\\tabucline[1.5pt]{-}
\end{tabu}
}}
\end{center}
\label{table9}
\end{table*}

\subsection{Tile-Level Zero-shot Classification Results}
We conducted tile-level zero-shot classification on five distinct datasets, evaluating only their test splits. The outcomes, detailed in Table \ref{table8}, benchmark our CPLIP algorithm against current SOTA VL-based methods across balanced accuracy, weighted $F_{1}$, and AUROC scores, all based on a single prompt. Comprehensive results using merged prompts are available in the supplementary material. Our CPLIP model consistently outperformed others in both single and merged prompt scenarios on all datasets. The CONCH algorithm was the next best, showing strong results on the CRC100K and SICAP datasets, though its performance on the other datasets was not documented.
%

CPLIP notably enhanced performance compared to CONCH, with gains of 13.5\% in balanced accuracy, 13.9\% in weighted $F_{1}$, and 2.1\% in AUROC for the CRC100K dataset using single prompts. For the SICAP dataset, the improvements were 1.7\% in balanced accuracy and 14.30\% in weighted $F_{1}$. Against MI-Zero/PLIP on the DigestPath and PanNuke datasets, CPLIP's enhancements were (1.3\%, 4.5\%, 2.2\%) and (2.2\%, 6.90\%, 3.0\%), respectively. On WSSS4LUAD, CPLIP outperformed MI-Zero by 5.6\% in balanced accuracy, 4.9\% in weighted $F_{1}$, and 3.1\% in AUROC. \textit{These significant performance improvements over SOTA methods demonstrate the advantages of our CPLIP algorithm.}

\subsection{WSI-Level Zero-shot Classification Results}
For zero-shot classification of gigapixel WSIs, we adopted an approach akin to MI-Zero \cite{lu2023visual}. 
We binarized each WSI to distinguish tissue from the background using the OTSU method and extracted $N$ number of tiles each with $224 \times 224$ pixels. Each tile's embedding was obtained via the CPLIP image encoder and $\ell_{2}$-normalization. We then calculated cosine similarities between tile embeddings and text embeddings, producing $C$ similarity scores per tile. These were aggregated using top-$K$ pooling, averaging the highest $K$  scores per class to determine the slide-level class prediction, with $K$ chosen from 1, 5, 10, 50, 100 based on best performance metrics (i.e., the highest balanced accuracy, weighted $F_{1}$, and AUROC scores for classification tasks).

Table \ref{table9} compares our CPLIP algorithm's zero-shot performance with SOTA VL models on CAM16, TCGA-BRCA, TCGA-RCC, and TCGA-NSCLC datasets, using a single prompt.
Detailed results with merged prompts are in the supplementary material.
CPLIP's performance was also assessed with various out-of-domain and in-domain encoders.
CPLIP consistently outperformed in-domain VL models like PLIP, BiomedCLIP, MI-Zero, and CONCH. 
For instance, CPLIP$_{2}$ in-domain zero-shot balanced accuracy reached 59.10\% for lymph node metastasis in CAM16, surpassing MI-Zero by 13.50\%. 
In NSCLC and RCC subtyping, CPLIP$_{2}$ achieved balanced accuracies of 85.40\% and 84.40\%, respectively, outperforming CONCH and MI-Zero by margins up to 5.20\%. 
Notably, in the BRCA subtyping task, CPLIP$_{2}$ achieved an 82.40\% balanced accuracy, significantly ahead of CONCH and MI-Zero by 18.10\% and 4.30\%, respectively. 
{\it These results highlight CPLIP$_{2}$ SOTA performance in cancer subtyping using zero-shot learning}.
\subsection{Zero-shot Segmentation of Gigapixel Images}
We also performed zero-shot slide-level segmentation similar to CONCH \cite{lu2023towards} using the SICAP (31 WSIs) and DigestPath (250 large images) datasets.
Overall, CPLIP outperformed other VL methods in both datasets by a significant margin demonstrating the advantages of heterogeneous textual descriptions and histology images. 
For further details, consult our supplementary material.

\section{Conclusion}
\label{sec:conclusion}
\vspace{-1mm}
Existing visual learning (VL) models in computational pathology require paired image and text data for zero-shot learning. 
In contrast, we propose an algorithm that enables unpaired alignment of image and textual data for zero-shot learning in histopathology. 
We construct a comprehensive bag of textual descriptions using heterogeneous sources including cancer glossaries, GPT-3, and off-the-shelf VL models. 
These are used to build a corresponding bag of visual concepts. 
A bag-based contrastive learning approach then aligns the textual and visual concepts semantically. 
Extensive experiments on nine independent datasets demonstrate the superior zero-shot classification and segmentation performance of our proposed Comprehensive Pathology Language Image Pre-training (CPLIP) algorithm compared to SOTA VL models. 
Our framework is inherently translational to other applications and, in the future, we aim to develop a comprehensive pathologyGPT model to enhance cancer diagnosis and prognostications.

\section{Acknowledgement}
\vspace{-1mm}
This publication acknowledges the support provided by the
Khalifa University under Faculty
Start-Up grants FSU-2022-003 Award No. 8474000401, and ASPIRE Grant: AARE20-279.

{
    \small

}

 
\section{Supplementary Material}

\begin{figure*}[t!]
\centering
\includegraphics[width=\linewidth]{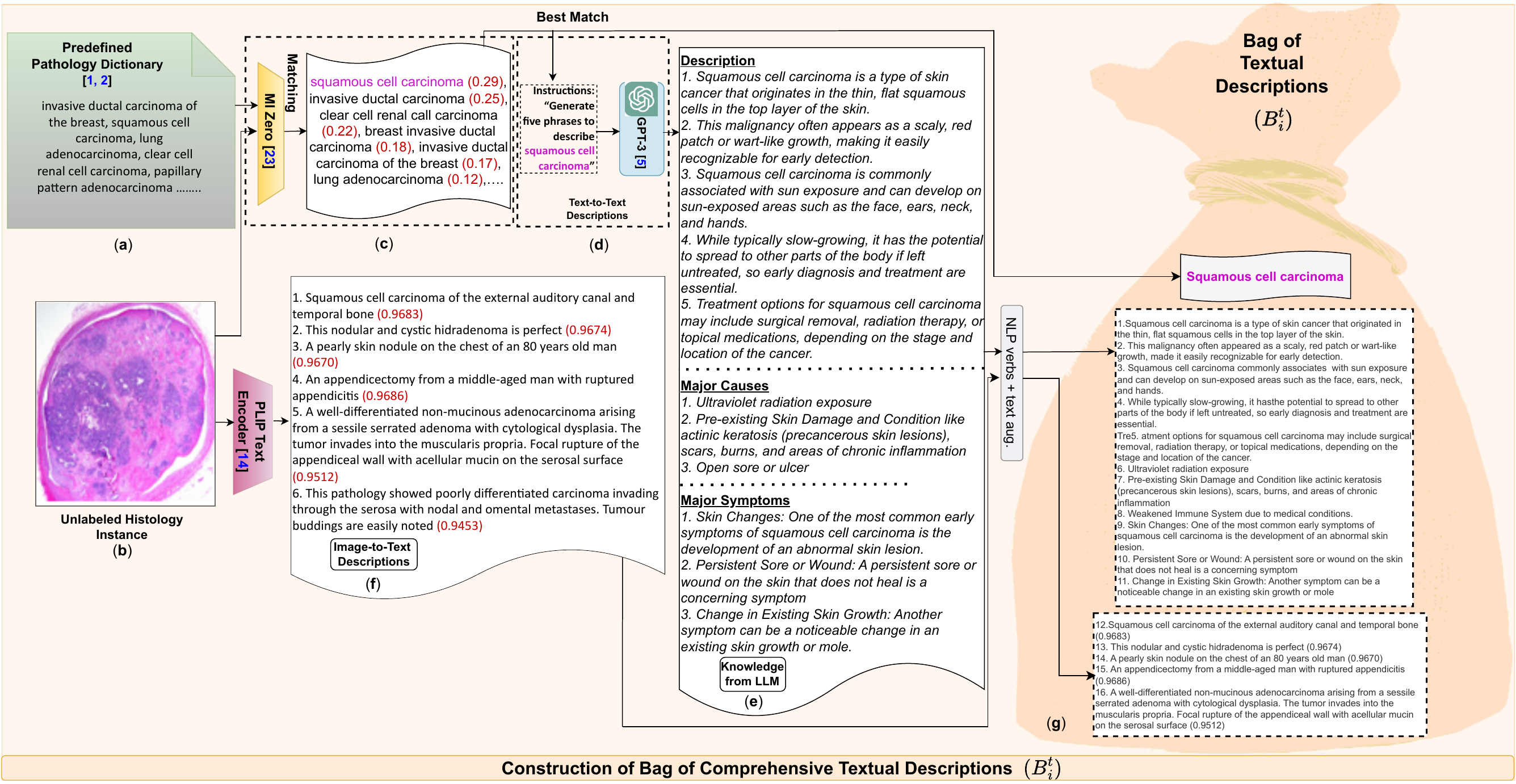}
\caption{Diagram outlining the {\bf detailed} construction process of the textual description bag ($B_{i}^{t}$) for best-matched prompt, \enquote{squamous cell carcinoma} shown in the main paper (Fig. \textcolor{blue}{3} (A)).
There are three primary steps: using MI-Zero to identify the best text match, leveraging GPT-3 to enrich the textual descriptions of the best-matched text, and employing the PLIP text encoder to generate more in-depth descriptions of the input unlabeled histology image.}
\label{fig2}
\end{figure*}

\subsection{Glossary} 
\begin{itemize}
    \item {\bf An open vocabulary} allows machine learning models to recognize and work with words they haven't encountered before, rather than being limited to a pre-set list of terms.
    \item {\bf Zero-shot transfer} refers to a machine learning model's ability to correctly handle tasks it has not been explicitly trained to perform, using knowledge learned during training from other related tasks.
    \item {\bf Paired image and text data} consist of sets of images each directly associated with descriptive text that explains or provides context for the visual content. This pairing is used to train models to understand and align the content and context between the visual and textual information.
    \item {\bf Whole Slide Images (WSIs)} are high-resolution digital scans of entire microscope slides containing tens of thousands of pixels used in pathology to examine tissues in detail.
    \\
    They are called {\bf Whole Slide Images (WSIs)} because they are comprehensive digital scans that capture the entire tissue sample present on a glass slide, typically used for pathological examination. This allows pathologists to view the slide in its entirety on a computer, zoom in on areas of interest, and perform detailed analyses that would traditionally be done under a microscope.
    \item {\bf Tile-level zero-shot learning} in the context of computational pathology refers to the ability of a machine learning model to classify individual tiles or patches of a whole slide image (WSI) into their correct categories without having been explicitly trained on those specific tiles or annotations. Each tile is a small, high-resolution section of a larger WSI, and the model must use learned patterns from other tasks or datasets to make accurate predictions.
    \item {\bf Cancer subtyping} is the process of classifying cancer into more specific categories based on its cellular characteristics, molecular profile, and behavior. This helps in understanding the prognosis and determining the most effective treatment approach for each specific type.
    \item {\bf Multi-Instance Learning (MIL)} is a variant of machine learning where data is grouped into 'bags' with a single label per bag, despite containing multiple instances. The MIL algorithm predicts bag labels by learning from the collective features of instances within each bag.
    \item {\bf Is Multi-Instance Learning (MIL) considered to be a type of weakly supervised learning?} Yes, because it deals with training data that has incomplete or ambiguous labels. In MIL, only the bag of instances is labeled, not the individual instances, which is a weaker form of supervision than having labels for every instance.
    \item {\bf MI-Zero} \cite{lu2023visual} is a framework designed to enhance the analysis of histopathology images, particularly gigapixel whole slide images used in medical diagnostics (see also main paper Sec. \textcolor{red}{3.2.1} for details). This framework is notable for its ``zero-shot transfer capabilities." These capabilities are derived from contrastively aligned image and text models, which are used to facilitate multiple cancer subtype classification tasks.
    \\
    Specifically, MI-Zero uses pre-trained encoders to analyze these complex histopathology images. The key advantage of this approach is that it does not require any additional labeling of the images, which can be a time-consuming and resource-intensive process in medical image analysis. By leveraging existing models and their zero-shot transfer capabilities, MI-Zero aims to streamline and improve the diagnostic process in histopathology, enhancing the efficiency and accuracy of analyses conducted on these detailed images.
    \item {\bf A lemma} is the base form of a word from which all its inflected or variant forms are derived. In the context of verbs, it's the form that appears in the dictionary, which is usually the present tense, singular form. For example, \enquote{go} is the lemma for \enquote{goes}, \enquote{going}, \enquote{went}, and \enquote{gone}. Lemmatization is the process of grouping together these different forms of a word so they can be analyzed as a single item. This is especially useful in natural language processing, where understanding the meaning of a word in different contexts is essential. {\bf This type of augmentation is likely used to improve the model's understanding by allowing it to recognize different forms of a verb as the same action or state}.
\end{itemize}

\begin{figure*}[t!]
\centering
\includegraphics[width=\linewidth]{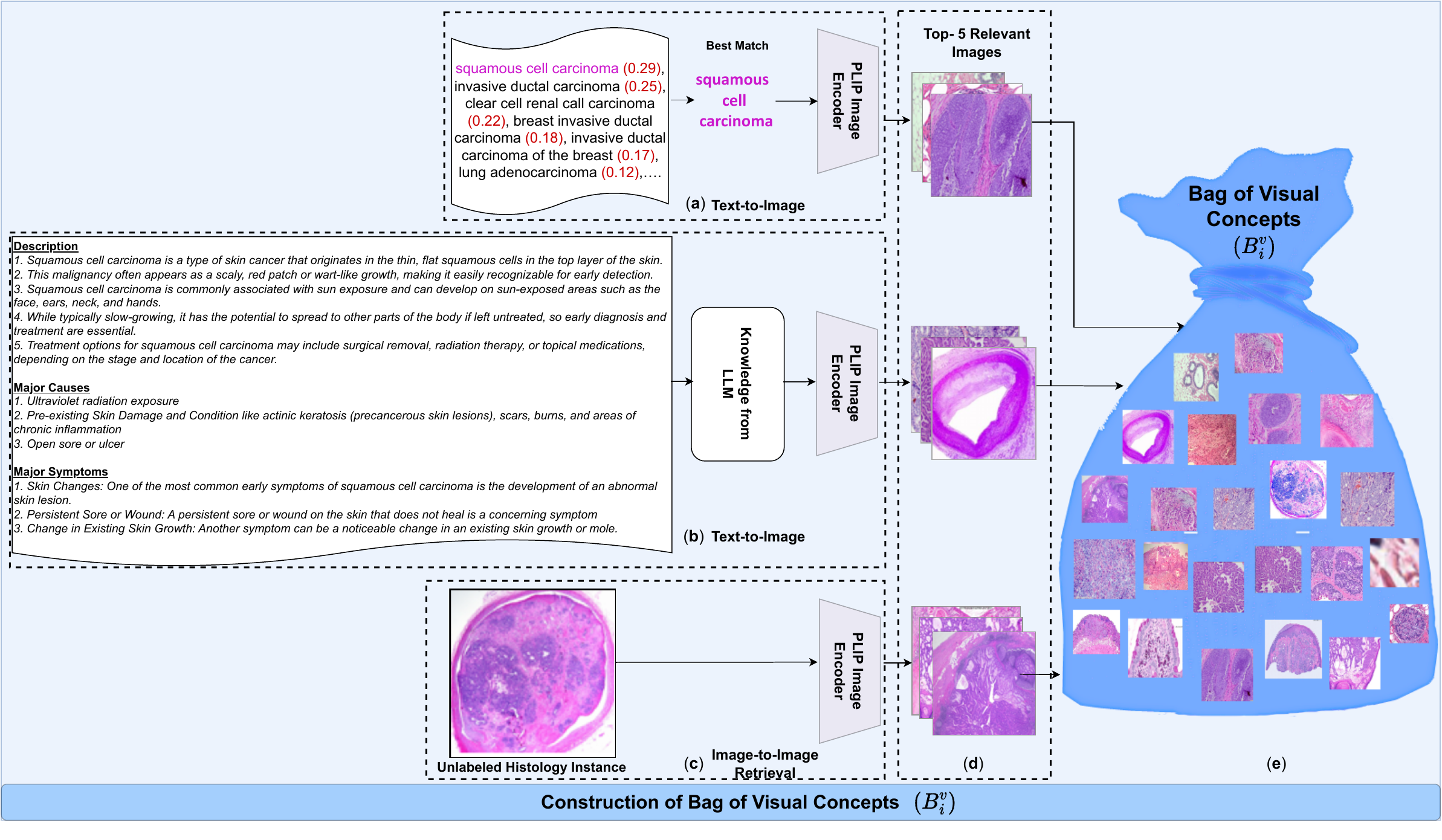}
\caption{This diagram details the steps taken to create the bag of visual concepts $B_{i}^{v}$ for the best-matched prompt \enquote{squamous cell carcinoma} shown in the main paper (Fig. \textcolor{blue}{3} (B)). 
The process involves {\bf (a)} using PLIP to select images that closely match the prompt, {\bf (b)} using PLIP to enrich the dataset with histology images that align with the best-matched textual descriptions, and {\bf (c)} employing PLIP to retrieve relevant histology images for the input unlabeled histology image}.
\label{fig3}
\end{figure*}
\subsection{Our CPLIP Model Overview, Context, and Insights}    
In computational pathology, vision-language models have shown their impact in classifying and analyzing WSIs for various tasks (see PLIP \cite{huang2023visual}, MI-Zero \cite{lu2023visual}, BiomedCLIP \cite{zhang2023large}, and CONCH \cite{lu2023towards}).
The textual cues are instrumental in optimizing the performance of VL models. 
However, the current models' reliance on a singular prompt for a given histology image may lead to potentially restricted performance for zero-shot classification \cite{huang2023visual, zhang2023large, lu2023visual}.
Typically, these models employ simple noun-based phrases like \enquote{\texttt{Photomicrograph showing clear cell change in oral squamous cell carcinoma}} or \enquote{\texttt{Photomicrograph of carcinomatous component (adenocarcinoma)}}, overlooking the causes and symptoms associated with specific cancer types (please see main paper Fig. \textcolor{red}{2} (a) for details). 
Integrating more descriptive prompts, such as \enquote{\texttt{squamous cell carcinoma is instigated by exposure to ultraviolet radiation and human papillomavirus}} and \enquote{\texttt{symptoms of squamous cell carcinoma include skin changes, persistent sore or wound, or changes in existing skin growth}}, could significantly enhance the information available to VL models during training (see Fig. \ref{fig2}). 

To our knowledge, no existing computational pathology VL models have incorporated such diverse textual prompts either during training or at the inference stage. 
Unlike existing methods focusing on aligning individual textual and visual concepts, we propose a simultaneous alignment of numerous interrelated textual and visual concepts (refer to Fig. \textcolor{red}{2} (b) in the main paper for details).

\begin{figure*}[t!]
\centering
\includegraphics[width=\linewidth]{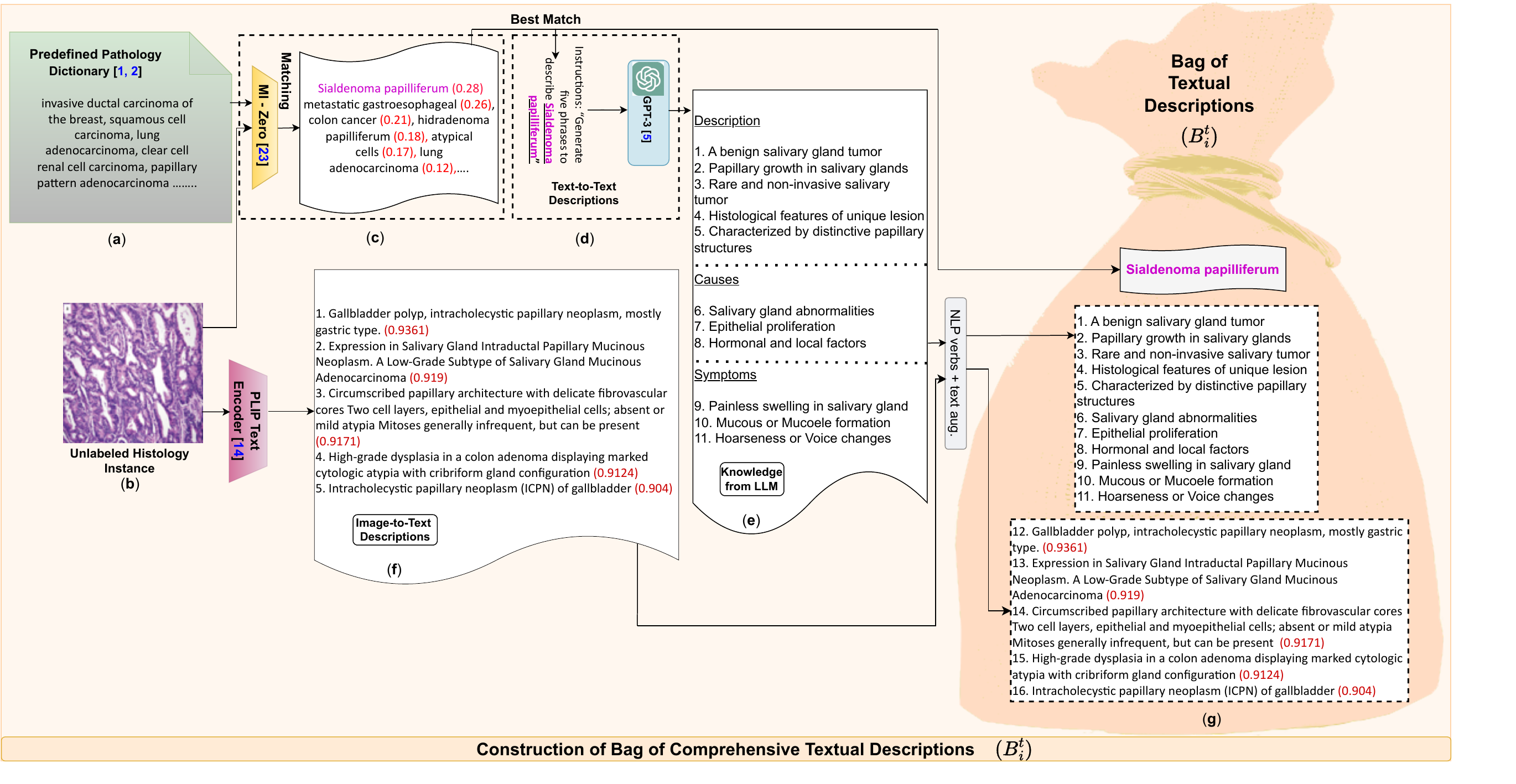}
\caption{This figure outlines the process for creating $B_{i}^{t}$ for the prompt \enquote{Sialdenoma papilliferum} as a second example. 
The procedure involves three primary steps: {\bf (a)} using MI-Zero to select the best text match,  {\bf (b)} enriching the textual descriptions with GPT-3 to add depth, and {\bf (c)} employing the PLIP text encoder to generate detailed descriptions of the input unlabeled histology image.}
\label{fig4}
\end{figure*}

We define ``comprehensiveness" as the incorporation of a broad array of textual descriptions for the same medical conditions, coupled with a diverse set of histology images for those conditions.
This approach acknowledges that a single disease may be described differently by various medical professionals and can manifest in multiple ways across patients. 
Despite these variances, combining different descriptions and images provides a holistic view, enhancing the VL models' ability to make connections between symptoms, causes, and specific medical conditions.
As shown in Figs. \ref{fig2} and \ref{fig4}, the best-matched prompt examples, \enquote{Squamous Cell Carcinoma} and \enquote{Sialdenoma papilliferum}, have a broad array of textual descriptions coupled with a diverse set of causes and symptoms.

We exploit this {\bf diversity} with a focus on \enquote{comprehensiveness}.
This term, in the context of textual prompts, refers to the variety of ways the same medical conditions (diseases) are described textually. Similarly, ``comprehensiveness" in visual concepts involves having numerous histology images for the same medical condition.
Our motivation is driven by the fact that medical practitioners often describe the same disease in various ways, and that the illness could manifest in varied forms for each patient. 
Despite these differences, the textual descriptions and the morphological characteristics of the disease are mutually informative. 
We propose to integrate detailed symptoms into the textual prompts, which would aid VL models in establishing correlations between symptoms, causes, and particular diseases or medical conditions. 
Moreover, we propose the integration of specific medical condition symptoms into the textual prompts, facilitating VL models in drawing correlations between symptoms, causes, and specific diseases or medical conditions.

To achieve comprehensive textual prompts, we initially compiled a dictionary of various cancer types and associated medical conditions by referencing multiple accessible online glossaries. 
Subsequently, for a given histology image, we assess the similarity and extract the most fitting prompts from the pathology dictionary using the existing VL model \cite{lu2023visual}. 
The best-matching prompt is then changed into five distinct variations using the GPT-3 model \cite{brown2020language}. 
Additionally, we identified three primary causes and three notable symptoms related to the same prompt, utilizing GPT-3. 
Along with the same histology image, we also ascertain the most appropriate textual descriptions and relevant tissue images from the Medical Twitter dataset by using the PLIP model.
Figs. \ref{fig2} and \ref{fig4} show two examples of best-matched prompts, \enquote{Squamous Cell Carcinoma} and \enquote{Sialdenoma papilliferum}, with textual descriptions using GPT-3 and PLIP.
To broaden the comprehensiveness/variety, we collect the most relevant tissue images through the PLIP model using the extended textual prompts obtained from GPT-3.
Figs. \ref{fig3} show the most relevant images associated with each textual description prompt analysed using the PLIP model.
We restrict the count of textual descriptions and histology images to 17 and 21, although the number of these items could be increased based on the available computational resources.

From the comprehensive textual prompts and visual concepts, we generate a bag of textual descriptions and a bag of images in an unsupervised and automated manner see two examples in Figs. \ref{fig2}-\ref{fig3}. 
Histology images corresponding to the same textual prompt in the established dictionary are classified as positive instances, while those corresponding to different prompts are labeled as negative instances. 
Using these comprehensive bags of textual and visual concepts, we fine-tune the baseline PLIP model to bring the embeddings of multiple positive textual and visual concepts closer together while distancing the embeddings of their negative counterparts. 
This process aims to boost class-agnostic representations (refer to the main paper, Fig. \textcolor{red}{1} (b)).

Our proposed fine-tuned model, termed the Comprehensive PLIP (CPLIP), can be employed in various downstream zero-shot classification tasks.
Through our proposed methodology, we have made progress towards enhancing the alignment between textual and visual embeddings by incorporating inclusive textual descriptions, which include disease symptoms and causes alongside multiple visual concepts. 
By minimizing a comprehensive contrastive loss function, we facilitate the alignment of multiple textual and visual elements, a strategy that has shown significant effectiveness in our experiments. This paves the way for future research in this field.
Building on this foundation, our proposed CPLIP model has demonstrated superior performance in a range of computational pathology tasks, including tile-based classification, WSI-level classification for cancer subtyping, and histology image segmentation in zero-shot settings without any further fine-tuning 
Our model demonstrated improved patch-based classification performance compared to existing SOTA methods on diverse datasets such as CRC100K \cite{kather2019predicting}, WSSS4LUAD \cite{han2022wsss4luad}, DigestPath \cite{da2022digestpath}, and PanNuke \cite{gamper2019pannuke}.
In addition, the proposed model has also obtained better results than the existing SOTA methods on TCGA-BRCA, TCGA-RCC, and TCGA-NSCLC datasets.

\subsection{Predefined Pathology Vocabulary}   
Table \ref{table_dictionary} shows our pathology prompts dictionary collected from online cancer glossaries \cite{cancer1, cancer2}.
Our pathology dictionary covers diverse cancer types and morphologies across various tissue types and includes terms commonly used by expert pathologists to describe various cancer forms, related medical conditions, and their prognoses through histology images.
This serves as a foundational prompt to match any input unlabelled histology images and extract comprehensive textual descriptions in subsequent phases.

%

\subsection{Training and Implementation Details}
In the field of histopathology, publicly accessible image-text paired datasets are scarce. 
The only publicly available image-text histology dataset is the ARCH dataset \cite{gamper2021multiple}, which consists of 8,617 image-text pairs derived from 12,676 journal articles on clinical and research pathology. 
We fine-tuned our proposed CPLIP algorithm using the ARCH dataset, which contains histology images without using their corresponding textual descriptions. 
We expanded each image into approximately 21 images and 17 different textual descriptions, resulting in a total of 180,000 images and 146,000 textual descriptions.
While the ARCH dataset was also used by MI-Zero \cite{lu2023visual} during training, they employed image-text paired data, whereas our algorithm requires unpaired many-to-many image-text alignment.

We fine-tuned our proposed CPLIP model by initializing weights using different image encoders and text encoders.
For a performance comparison of our CPLIP model under different settings, refer to Table \ref{ablation2}. 
The following different image-text encoders were employed during fine-tuning our CPLIP model:


\begin{enumerate}
\item Similar to the other SOTA methods \cite{huang2023visual, zhang2023large, lu2023visual}, we fine-tuned the baseline CLIP \cite{radford2021learning} with a ViT-B/16-224 \cite{ViT-16} as image encoder and a GPT-2/77 \cite{radford2019language} as text encoder (Table \ref{ablation2}).
\item Given that the baseline CLIP is trained on out-of-domain paired data, we also fine-tuned our CPLIP model using a pathology domain-specific pre-trained PLIP \cite{huang2023visual} with a PLIP-ViT-B/32-224 as image encoder and a GPT-2/347 as text encoder (Table \ref{ablation2}).
\item Additionally, we fine-tuned the CPLIP model using BioClinicalBert/512 \cite{alsentzer2019publicly} and PubMedBERT/256 \cite{gu2021domain} as the text encoders and CTransPath/224 \cite{wang2022transformer} as the image encoder, similar to MI-zero and BiomedCLIP \cite{zhang2023large} (Table \ref{ablation2}). 
BioClinicalBert and PubMedBERT are medical-specific non-pathology text encoders trained on biomedical and clinical corpora such as PubMed abstracts and MIMIC \cite{johnson2016mimic}, while CTransPath is trained using self-supervised representation learning on a total of 15.5 million unlabeled histology patches. 
Both of these encoders utilized ViT-B/16.
\item We fine-tuned our CPLIP algorithm using BioClinicalBert/512 as the text encoder and PLIP-ViT-B/32-224 as the in-domain image encoder (Table \ref{ablation2}). 
\item We also fine-tuned our CPLIP algorithm using CTransPath/224 as the in-domain image encoder and PLIP-GPT/347 as the in-domain text encoder (Table \ref{ablation2}).
\end{enumerate}

\noindent Across all visual-language pre-training variants, we trained our models using a temperature parameter of 0.02, the AdamW optimizer \cite{ICLRADAM} with an initial learning rate of $5 \times 10^{-6}$, and a cosine decay scheduler. 
We trained our models with a 
batch size of 256 for 50 epochs. 
We set the filtering thresholds $\delta_{t}$ and $\delta_{v}$ to discard 10\% of the data from each bag. 
After filtering, the bag of textual descriptions was reduced to 15 items and the bag of visual concepts was reduced to 19 items. 
Experiments are conducted using both single and merged prompts at inference time similar to \cite{lu2023towards} for fair comparison.


\subsection{Datasets}
\noindent \textbf{1 .CRC100K \cite{kather2019predicting}:} is a colorectal cancer dataset containing $224 \times 224$ pixels tiles captured at 0.5 microns per pixel extracted from 50 patients.
The dataset contains nine distinct tissue types including colorectal adenocarcinoma epithelium,  normal colon mucosa, smooth muscle, lymphocytes, mucus, cancer-associated stroma, adipose, background, and debris.
The official training (100K images) and testing (7,180 images) splits are provided.
For zero-shot tile-based classification, we used the testing split without any fine-tuning.

\noindent  \textbf{2. WSSS4LUAD \cite{han2022wsss4luad}:} is a lung adenocarcinoma dataset containing tiles with almost $200 \times 500$ pixels.
The dataset contains three distinct classes: tumor, tumor-associated stroma, and/or normal. 
Similar to PLIP, we performed binary classification of Tumor Vs. Normal.
The training dataset contains 7063 images while the testing data consists of 3028 images (2015 Tumor, 1013 Normal).
For zero-shot tile-based classification, we used the testing split without any fine-tuning.

\noindent  \textbf{3. SICAP \cite{silva2021self}:} is a prostate cancer dataset for Gleason pattern classification consisting of $512\times 512$ pixels tiles extracted from 155 WSIs.
The official training split consists of 9,959 images from 124 WSIs and the testing split consists of 2,122 images from 31 WSIs.
The dataset contains four labels as the primary Gleason pattern (3, 4, or 5) or as non-cancerous (NC). 
We employed the official testing split for zero-shot classification experiments.

\noindent  \textbf{4. PanNuke \cite{gamper2019pannuke}:} is a more diverse nuclei segmentation and classification dataset consisting of 19 different tissue types.
The training and testing splits consist of 4346 and 1888 images with $256 \times 256$ pixels.
Similar to the PLIP, we evaluate the zero-shot classification performance of the proposed algorithm for Tumor vs. Normal Benign classes using the testing split.

\noindent  \textbf{5. DigestPath \cite{da2022digestpath}:} is a dataset of colonoscopy H \& E tissue sections consisting of 660 images. 
Similar to PLIP, we performed tile-based zero-shot classification for Tumor Vs. Normal on the testing split containing 18814 images.
For zero-shot segmentation, we employed the official 250 images from 93 patients for which pixel-level lesion annotation for colorectal cancer tissue is provided for testing.

\noindent  \textbf{6. CAM16 \cite{bejnordi2017diagnostic}:}  is a breast cancer dataset for lymph node metastasis detection using gigapixel WSIs.
The total number of WSIs is 400 with only slide-level labels are provided.
The official training split contains 270 WSIs and the testing split contains 130 testing WSIs.
In training, the total number of normal WSIs is 159, and that containing tumor regions of breast cancer metastasis is 111.
For zero-shot WSI-level classification, we used only the official testing split.

\noindent  \textbf{7. TCGA-BRCA\footnote {portal.gdc.cancer.gov \label{f1}} :} is a TCGA dataset of invasive breast carcinoma containing two types of WSIs including Invasive Ductal Carcinoma (IDC) and Invasive Lobular Carcinoma (ILC).
The total number of WSIs is 1048 of which 837 are IDC and 211 are ILC.
For zero-shot WSI-level classification, similar to CONCH, the test set consists of 75 WSIs from each class with no patient-level overlap between the training and testing splits.

\noindent  \textbf{8. TCGA-RCC$^{\ref{f1}}$:} is a TCGA dataset of renal cell carcinoma containing three types of WSIs including Clear Cell Renal Cell Carcinoma (CCRCC), Papillary Renal
Cell Carcinoma (PRCC), and Chromophobe Renal Cell Carcinoma (CHRCC).
The total number of WSIs is 922 of which 519 are CCRCC, 294 are PRCC, and 109 are CHRCC.
For zero-shot WSI-level classification, similar to the CONCH, the test set consists of 75 WSIs from each of the three classes.
There is no patient-level overlap between the training and testing splits.

\noindent  \textbf{9. TCGA-NSCLC$^{\ref{f1}}$:}  is a TCGA dataset of Non-Small Cell Lung Cancer (NSCLC) containing two types of WSIs including LUng AaDenocarcinoma (LUAD) and LUng Squamous
cell Carcinoma (LUSC) cases.
The total number of WSIs is 1041 of which 529 are LUAD and 512 are LUSC.
For zero-shot WSI-level classification, similar to CONCH, the test set consists of 75 WSIs from each of the two classes.
There is no patient-level overlap between the training and testing splits.
\subsection {Evaluation Metrics}
We employed different evaluation metrics to evaluate the performance classification and segmentation tasks.
For the classification task, we employed balanced accuracy, weighted $F_{1}$ score, and AUCROC.
Balanced accuracy is defined as the macro average of the recall of each class. 
The weighted $F_{1}$ score is computed by taking the average of the $F_{1}$ score (the harmonic mean of precision and recall) of each class, weighted by the support of each class. 
In the binary case, AUCROC is the area under the receiver operating curve, which plots the true positive rate against the false positive rate as the classification threshold is varied. 
AUCROC is generalized to the multi-class case by averaging over the AUCROC of all pairwise combinations of classes.
For the segmentation task, we report the Dice score, which is the same as the $F_{1}$ score, and the precision and recall of the positive class.
The same set of evaluation metrics are also used by recent SOTA computational pathology VL models \cite{huang2023visual, lu2023towards}.


\begin{table}[t!]
\caption{Ablation 1: Zero-shot classification performance comparison in terms of weighted average $F_{1}$ score using single vs. merged prompts.
Significant performance improvement is observed using the merged prompts.95\% Confidence Interval (CI) is included in parentheses.
}
\begin{center}
\makebox[\linewidth]{
\scalebox{0.92}{
\begin{tabu}{|c|c|c|}
\tabucline[0.5pt]{-}
Ablation Study&Single Prompts&Merged Prompts\\\tabucline[0.5pt]{-}
CRC100K&
\underline{0.681}(
\underline{0.663},
\underline{0.702})&
\textbf{0.844}(
\textbf{0.833},
\textbf{0.856})\\\tabucline[0.5pt]{-}
DigestPath&
\underline{0.856}(
\underline{0.875},
\underline{0.889})&
\textbf{0.903}(
\textbf{0.891},
\textbf{0.915})\\\tabucline[0.5pt]{-}
SICAP&
\underline{0.388}(
\underline{0.375},
\underline{0.395})&
\textbf{0.511}(
\textbf{0.498},
\textbf{0.526})
\\\tabucline[0.5pt]{-}
WSSS4LUAD&
\underline{0.791}(
\underline{0.784},
\underline{0.805})&
\textbf{0.882}(
\textbf{0.876},
\textbf{0.894})\\\tabucline[0.5pt]{-}
PanNuke&
\underline{0.757}(
\underline{0.741},
\underline{0.763})&
\textbf{0.811}(
\textbf{0.799},
\textbf{0.827})
\\\tabucline[0.5pt]{-}
\end{tabu}
}}
\end{center}
\label{ablation1}
\end{table}

\begin{table*}[t!]
\caption{Ablation 2: Zero-shot classification performance comparison in terms of weighted average $F_{1}$ score using different pre-trained image and text encoders. 
95\% Confidence Interval (CI) is included in parentheses.
All experiments use ViT-B/16 as the image encoder and PubMedBERT or BioClinicalBert to initialize the text encoder. 
Please note the performance is reported using merged prompts.}
\begin{center}
\makebox[\linewidth]{
\scalebox{0.80}{
\begin{tabu}{|c|c|c|c|c|c|c|c|}
\tabucline[0.5pt]{-}
Ablation Study&Vision Encoder&Text Encoder&CRC100K&DigestPath&SICAP&WSSS4LUAD&PanNuke\\\tabucline[0.5pt]{-}
CPLIP&CLIP&CLIP&0.611&0.803&0.344&0.765&0.708\\
(Out-of-domain)&(ViT-B/16-224)&(GPT-2/77)&(0.588,0.634)&(0.794, 0.812)&(0.305,0.383)&(0.731,0.796)&(0.692,0.714)\\\tabucline[0.5pt]{-}
CPLIP&PLIP&PLIP&0.828&0.886&0.502&0.804&0.802\\
(In-domain)&(ViT-B/32-224)&(GPT/347)&(0.802,0.841)&(0.873, 0.804)&(0.491,0.511)&(0.791,0.815)&(0.793,0.814)\\\tabucline[0.5pt]{-}
\textbf{CPLIP}&\textbf{CTransPath}&\textbf{BioClinicalBert} &\textbf{0.844}&\textbf{0.903}&\textbf{0.511}&\textbf{0.882}&\textbf{0.811}\\
(\textbf{Out-of-domain})&(\textbf{ViT-B/16-224})&(\textbf{BioClinicalBert/512})&(\textbf{0.833},\textbf{0.856})&(\textbf{0.891},\textbf{0.915})&(\textbf{0.498},\textbf{0.526})&(\textbf{0.876},\textbf{0.894})&(\textbf{0.799},\textbf{0.827})\\\tabucline[0.5pt]{-}
CPLIP&CTransPath&PubMedBERT&\underline{0.838}&\underline{0.894}&\underline{0.508}&\underline{0.866}&\underline{0.807}\\
(Out-of-domain)&(ViT-B/16-224)&(PubMedBERT/256)&(\underline{0.828},\underline{0.847})&(\underline{0.885},\underline{0.905})&(\underline{0.491},\underline{0.518})&(\underline{0.863},\underline{0.881})&(\underline{0.793},\underline{0.819})\\\tabucline[0.5pt]{-}
CPLIP&PLIP&PubMedBERT&0.825&0.881&0.482&0.841&0.782\\
(Out-of-domain)&(ViT-B/32-224)&(PubMedBERT/256)&(0.804,0.874)&(0.854,0.913)&(0.441,0.517)&(0.822,0.863)&(0.756,0.815)\\\tabucline[0.5pt]{-}
CPLIP&PLIP&BioClinicalBert&0.828&0.891&0.494&0.871&0.798\\
(Out-of-domain)&(ViT-B/32-224)&(BioClinicalBert/512)&(0.811,0.840)&(0.880,0.905)&(0.455,0.531)&(0.851,0.891)&(0.766,0.823)\\\tabucline[0.5pt]{-}
CPLIP&CTransPath&PLIP&0.831&0.892&0.496&0.844&0.777\\
(In-domain)&(ViT-B/16-224)&(GPT/347)&(0.821,0.843)&(0.881,0.835)&(0.471,0.512)&(0.822,0.867)&(0.761,0.786)\\\tabucline[0.5pt]{-}
\end{tabu}
}}
\end{center}
\label{ablation2}
\end{table*}

\begin{table*}[t!]
\caption{Ablation 3: Zero-shot classification performance comparison in terms of weighted average $F_{1}$ score for varying size of text bag ($\delta_{t}=100\%$). 
95\% Confidence Interval (CI) is included in parentheses. 
Please note the performance is reported using merged prompts.}
\begin{center}
\makebox[\linewidth]{
\scalebox{0.90}{
\begin{tabu}{|c|c|c|c|c|c|}
\tabucline[0.5pt]{-}
Bag size ($B^{t}$)&1&5&10&15&17\\\tabucline[0.5pt]{-}
CRC100K&0.766(0.751,0.788)&0.788(0.751,0.812)&0.811(0.803,0.827)&\textbf{0.844}(\textbf{0.833},\textbf{0.856})&
\underline{0.841}(
\underline{0.821},
\underline{0.861})\\\tabucline[0.5pt]{-}
DigestPath&0.856(0.825, 0.882)&0.881(0.852, 913)&0.901(0.871, 0.9410)&\textbf{0.903}(\textbf{0.891}, \textbf{0.915})&
\underline{0.902}(
\underline{0.895},
\underline{0.916})\\\tabucline[0.5pt]{-}
WSSS4LUAD&0.841(0.832,0.856)&0.856(0.846,0.867)&0.875(0.861,0.889)&\textbf{0.882}(\textbf{0.876},\textbf{0.894})&
\underline{0.881}(
\underline{0.856},
\underline{0.916})\\\tabucline[0.5pt]{-}
SICAP&0.401(0.360,0.443)&0.433(0.413,0.466)&0.471(0.453,0.498)&\textbf{0.511}(\textbf{0.498},\textbf{0.526})&
\underline{0.499}(
\underline{0.472},
\underline{0.521})\\\tabucline[0.5pt]{-}
PanNuke&0.766(0.733,0.792)&0.781(0.752,0.812)&0.803(0.791,0.811)&\textbf{0.811}(\textbf{0.799},\textbf{0.827})&
\underline{0.815}(
\underline{0.791},
\underline{0.827})\\\tabucline[0.5pt]{-}
\end{tabu}
}}
\end{center}
\label{ablation3}
\end{table*}

\begin{table}[t!]
\caption{Zero-shot classification performance, in terms of weighted average $F_{1}$ score, is compared between two contrastive learning approaches: one-to-one (CPLIP$_{o}$) and many-to-many (CLIP), using a single prompt. 
Significant improvements in performance are seen with the use of the proposed many-to-many contrastive learning method, as demonstrated across four datasets.}
\begin{center}
\makebox[\linewidth]{
\scalebox{0.85}{
\begin{tabu}{|c|c|c|}
\tabucline[0.5pt]{-}
Datasets&One-to-One (CPLIP$_{o}$)&Many-to-Many (CLIP)\\\tabucline[0.5pt]{-}
CRC100K&\underline{0.656}&\textbf{0.681}\\\tabucline[0.5pt]{-}
SICAP&\underline{0.341}&\textbf{0.388}\\\tabucline[0.5pt]{-}
TCGA-BRCA&\underline{0.732}&\textbf{0.786}\\\tabucline[0.5pt]{-}
TCGA-RCC&\underline{0.821}&\textbf{0.855}\\\tabucline[0.5pt]{-}
\end{tabu}
}}
\end{center}
\label{table_many}
\end{table}

\subsection{Ablation Studies}

\noindent \textbf{1. Zero-shot performance comparison using single vs. merged prompts (Table \ref{ablation1}).}
In this experiment, we compared the zero-shot classification performance of the proposed CPLIP algorithm using single prompts vs. merged prompts at the inference step (see Table \ref{ablation1}).
For a fair comparison with earlier works \cite{huang2023visual, lu2023visual, zhang2023large}, we have used the same set of merged prompts as employed by CONCH \cite{lu2023towards}. 
On all five datasets for tile-based zero-shot classification, significant performance improvement is observed which is in line with the previous studies \cite{huang2023visual, lu2023towards}. 
\\

\noindent \textbf{2. Zero-shot performance comparison using different image-text encoders (Table \ref{ablation2}).}
In this experiment, we compared the performance of our proposed CPLIP algorithm in terms of initializing different image-text encoders including CLIP (out-of-domain pre-trained encoders), PLIP (in-domain pre-trained encoders), CTransPath (in-domain pre-trained image encoder), BioclinicalBert and PubMedBERT (out-of-domain pre-trained text encoders) as shown in Table \ref{ablation2}.
The best results on five datasets are reported using CTransPath as an image encoder and BioClinicalBert to initialize the text encoder.
This is because CTransPath is pre-trained on unlabeled larger histology images and BioClinicalBert is trained on 2M clinical notes in the MIMIC-III v1.4 database \cite{johnson2016mimic}.
The in-domain CPLIP variants also showed comparable performance compared to the best-performing CPLIP (out-of-domain) variant.
\\

\noindent \textbf{3. Zero-shot performance comparison using different sizes of bags (Table \ref{ablation3}).}
Experiments are also performed by varying the sizes of both bags.
In the textual bag ($B^{t}$), rank-1 best-matching textual description with the input image, rank-5, rank-10, rank-15, and all 17 textual descriptions are considered.
The corresponding visual bags ($B^{v}$) also contain 1, 5, 10, 15, and 21 images.
As the bag sizes increase, continuous improvements in performance are observed until bag size 15, as shown in Table \ref{ablation3}.
A further increase has caused a slight decrease in performance due to noisy textual descriptions.
\\

\noindent \textbf{4. Many-to-Many Vs. One-to-One Contrastive Learning Approach (Table \ref{table_many}).}
The many-to-many learning approach has two main advantages: {\bf (1)} it better reflects actual medical practice. 
Pathologists evaluate WSIs using not just visual morphology, but also patient symptoms and medical knowledge about disease causes.
By integrating these multiple data sources, the approach allows for more comprehensive clinical integration compared to previous one-to-one methods; {\bf (2)} the approach enhances visual representations through augmented slide image inputs, capturing phenotypic diversity and improving model generalization.
This methodology constructs comprehensive textual descriptions and corresponding visual concepts to enable VLMs to handle the complexity of pathology images and text. 
The approach is akin to multi-task learning as learning joint representations across related tasks can promote generalization - analogous to how multi-task learning leads to more robust models.
To assess its efficacy, an ablation study (Table \ref{table_many}) was conducted, revealing that this novel strategy significantly boosts zero-shot learning performance on four different datasets. 
Additionally, when compared to four SOTA vision-language models fine-tuned on the same datasets (Table \ref{table_data}), the many-to-many contrastive learning-based CPLIP model demonstrated superior accuracy, highlighting the robustness of this new technique.
\\

\noindent \textbf{5. Comparing CPLIP with in-domain SSL single modality CNNs and ViTs:}
Table \ref{table_image} compares the classification performance of CPLIP against domain-specific SSL CNNs and ViTs, namely DinoSSLPath \cite{kang2023benchmarking} and MoCo v2 \cite{chen2020improved}.
The study spanned four datasets at both WSI and tile levels, using zero-shot, linear evaluation, and full fine-tuning evaluation protocols.
Notably, zero-shot evaluations using CPLIP with merged prompts surpassed the performance of DinoSSLPath and MoCo v2, which did not employ zero-shot settings. 
In linear and fine-tuning settings, CPLIP achieved significant performance gains, maintaining consistency with the protocols established by DinoSSLPath.

\begin{table}[t!]
\caption{The zero-shot classification performance of SOTA methods, evaluated using a single prompt in terms of the weighted average $F_{1}$ score on data generated by our proposed approach, shows significant improvements across all SOTA models. 
CPLIP stands out as the top performer.}
\begin{center}
\makebox[\linewidth]{
\scalebox{0.73}{
\begin{tabu}{|c|c|c|c|c|}
\tabucline[0.5pt]{-}
Datasets&TCGA-NSCLC&TCGA-RCC&WSSS4LUAD&DigestPath\\\tabucline[0.5pt]{-}
CLIP&0.488&0.291&0.541&0.137\\\tabucline[0.5pt]{-}
BiomedCLIP&0.733&0.714&0.571&0.671\\\tabucline[0.5pt]{-}
PLIP&0.744&0.751&0.761&\underline{0.842}\\\tabucline[0.5pt]{-}
MI-Zero&\underline{0.811}&\underline{0.815}&\underline{0.762}&0.823\\\tabucline[0.5pt]{-}
CPLIP&\textbf{0.835}&\textbf{0.855}&\textbf{0.791}&\textbf{0.856}\\\tabucline[0.5pt]{-}
\end{tabu}
}}
\end{center}
\label{table_data}
\end{table}

\begin{table}[t!]
\caption{Comparative  classification  performance in terms of the weighted average $F_{1}$ score using 3 methods: zero-shot learning, linear evaluation, and fine-tuning of the proposed CPLIP, which uses merged prompts vs. DinoSSLPath and MoCo v2 (with ResNet50). 
Significant performance improvements are observed in case of linear evaluation and full fine-tuning of CPLIP.}
\vspace{-7mm}
\begin{center}
\makebox[\linewidth]{
\scalebox{0.80}{
\begin{tabu}{|c|c|c|c|c|}
\tabucline[1.0pt]{-}
Datasets&Evaluation&DinoSSLPath&MoCo v2&CPLIP\\\tabucline[1.0pt]{-}
\multirow{ 3}{*}{CAM16}&Zero-shot&$\times$&$\times$&\textbf{0.632}\\\tabucline[0.5pt]{2-5}
&Linear&\underline{0.618}&0.592&\textbf{0.663}\\\tabucline[0.5pt]{2-5}
(WSI-level)&Fine-tune&\underline{0.722}&0.678&\textbf{0.746}\\\tabucline[1.0pt]{-}
\multirow{ 3}{*}{WSSS4LUAD}&Zero-shot&$\times$&$\times$&\textbf{0.882}\\\tabucline[0.5pt]{2-5}
&Linear&0.878&\underline{0.881}&\textbf{0.924}\\\tabucline[0.5pt]{2-5}
(Tile-based)&Fine-tune&\underline{0.951}&0.944&\textbf{0.976}\\\tabucline[1.0pt]{-}
\multirow{ 3}{*}{CRC100K}&Zero-shot&$\times$&$\times$&\textbf{0.844}\\\tabucline[0.5pt]{2-5}
&Linear&\underline{0.862}&0.853&\textbf{0.894}\\\tabucline[0.5pt]{2-5}
(Tile-based)&Fine-tune&\underline{0.945}&0.911&\textbf{0.964}\\\tabucline[1.0pt]{-}
\multirow{ 3}{*}{SICAP}&Zero-shot&$\times$&$\times$&\textbf{0.511}\\\tabucline[0.5pt]{2-5}
&Linear&\underline{0.502}&0.466&\textbf{0.554}\\\tabucline[0.5pt]{2-5}
(Tile-based)&Fine-tune&\underline{0.604}&0.547&\textbf{0.626}\\\tabucline[1.0pt]{-}
\end{tabu}
}}
\end{center}
\label{table_image}
\end{table}

\begin{table*}[t!]
\caption{Tile-level zero-shot classification performance comparison in terms of balanced accuracy, weighted $F_{1}$, and AUCROC scores with existing VL-based models in computational pathology on five independent external datasets. On the WSSS4LUAD dataset, CONCH used a different split for performance evaluation which is indicated by $^{*}$. CPLIP performance is reported using the best combination from ablation study 2 (Table. \ref{ablation2}).}
\begin{center}
\makebox[\linewidth]{
\scalebox{0.90}{
\begin{tabu}{|c|c|c|c|c|c|}
\tabucline[1.5pt]{-}
Single Prompt&CRC100K&DigestPath&SICAP&WSSS4LUAD&PanNuke\\\tabucline[1.5pt]{-}
CLIP baseline \cite{radford2021learning}&0.234$|$0.185$|$0.727&0.11$|$0.030$|$0.203&0.231$|$0.139$|$0.201&0.451$|$0.481$|$0.705&0.322$|$0.352$|$0.683\\\tabucline[0.5pt]{-}
BiomedCLIP \cite{zhang2023large}&0.422$|$0.372$|$0.859&0.591$|$0.622$|$0.781&\textbf{0.381}$|$\underline{0.361}$|$0.506&0.466$|$0.495$|$0.698&0.522$|$0.572$|$0.711\\\tabucline[0.5pt]{-}
PLIP \cite{huang2023visual}&0.520$|$0.517$|$0.879&0.815$|$\underline{0.832}$|$0.901&0.319$|$0.255$|$0.603&0.702$|$0.734$|$\underline{0.822}&0.629$|$0.656$|$\underline{0.805}\\\tabucline[0.5pt]{-}
MI-Zero \cite{lu2023visual}&0.544$|$0.536$|$0.872&\underline{0.822}$|$0.811$|$\underline{0.911}&0.308$|$0.251$|$\underline{0.605}&\underline{0.722}$|$\underline{0.742}$|$0.805&\underline{0.659}$|$\underline{0.688}$|$0.755\\\tabucline[0.5pt]{-}
CONCH \cite{lu2023towards}&\underline{0.566}$|$\underline{0.542}$|$\underline{0.901}&-&0.349$|$0.245$|$-&0.598$^{*}$$|$0.590$^{*}$$|$0.795$^{*}$&-\\\tabucline[0.5pt]{-}
Proposed CPLIP &\textbf{0.701}$|$\textbf{0.681}$|$\textbf{0.922}&\textbf{0.835}$|$\textbf{0.856}$|$\textbf{0.933}&\underline{0.366}$|$\textbf{0.388}$|$\textbf{0.711}&\textbf{0.778}$|$\textbf{0.791}$|$\textbf{0.836}&\textbf{0.681}$|$\textbf{0.757}$|$\textbf{0.835}\\\tabucline[1.5pt]{-}
Merged Prompts&CRC100K&DigestPath&SICAP&WSSS4LUAD&PanNuke\\\tabucline[1.5pt]{-}
CLIP baseline \cite{radford2021learning}&0.271$|$0.247$|$0.781&0.188$|$0.210$|$0.280&0.283$|$0.191$|$0.205&0.501$|$0.544$|$0.791&0.385$|$0.412$|$0.744\\\tabucline[0.5pt]{-}
BiomedCLIP \cite{zhang2023large}&0.553$|$0.533$|$0.924&0.644$|$0.671$|$0.831&\underline{0.483}$|$0.439$|$0.605&0.511$|$0.533$|$0.764&0.631$|$0.651$|$0.802\\\tabucline[0.5pt]{-}
PLIP \cite{huang2023visual}&0.674$|$0.687$|$0.944&\underline{0.865}$|$\underline{0.871}$|$0.931&0.355$|$0.315$|$\underline{0.656}&\underline{0.751}$|$\underline{0.791}$|$0.833&0.719$|$0.744$|$0.874\\\tabucline[0.5pt]{-}
MI-Zero \cite{lu2023visual}&0.721$|$0.755$|$0.956&0.844$|$0.866$|$\underline{0.941}&0.341$|$0.306$|$0.641&0.741$|$0.781$|$\underline{0.846}&\underline{0.744}$|$\underline{0.759}$|$\underline{0.901}\\\tabucline[0.5pt]{-}
CONCH \cite{lu2023towards}&\underline{0.791}$|$\underline{0.803}$|$\underline{0.979}&-&\textbf{0.624}$|$0.424$|$-&0.719$^{*}$$|$0.705$^{*}$$|$0.877$^{*}$&-\\\tabucline[0.5pt]{-}
Proposed CPLIP &\textbf{0.823}$|$\textbf{0.844}$|$\textbf{0.980}&\textbf{0.871}$|$\textbf{0.903}$|$\textbf{0.971}&\underline{0.498}$|$\textbf{0.511}$|$\textbf{0.716}&\textbf{0.851}$|$\textbf{0.882}$|$\textbf{0.903}&\textbf{0.795}$|$\textbf{0.811}$|$\textbf{0.936}\\\tabucline[0.5pt]{-}
\end{tabu}
}}
\end{center}
\label{table8}
\end{table*}

\begin{table*}[t!]
\caption{WSI-level zero-shot classification performance comparison in terms of balanced accuracy, weighted $F_{1}$, and AUCROC scores with existing VL-based models in computational pathology on five independent external datasets. On the WSSS4LUAD dataset, CONCH used a different split for performance evaluation which is indicated by $^{*}$. We employed similar merged prompts during inference as proposed in CONCH \cite{lu2023towards}. (OoD: Out-of-Domain, InD: In-Domain)}
\begin{center}
\makebox[\linewidth]{
\scalebox{0.70}{
\begin{tabu}{|c|c|c|c|c|c|c|}
\tabucline[1.5pt]{-}
Models (Single prompts)&Image encoder pretraining&Text encoder pretraining&CAM16&TCGA-BRCA&TCGA-RCC&TCGA-NSCLC\\\tabucline[1.5pt]{-}
CLIP baseline \cite{radford2021learning}&ViT-B/16-224&GPT-2/77&0.134$|$0.175$|$0.325&0.512$|$0.328$|$0.551&0.321$|$0.178$|$0.578&0.496$|$0.358$|$0.536\\\tabucline[0.5pt]{-}
BiomedCLIP \cite{zhang2023large}&ViT-B/16-224&PMB/256&0.311$|$0.377$|$0.545&0.527$|$0.422$|$0.761&0.677$|$0.646$|$0.872&0.699$|$0.684$|$0.851\\\tabucline[0.5pt]{-}
PLIP \cite{huang2023visual}&ViT-B/32-224&GPT/347&0.399$|$0.416$|$0.681&0.451$|$0.331$|$0.611&0.726$|$0.739$|$0.915&0.676$|$0.666$|$0.781\\\tabucline[0.5pt]{-}
MI-Zero \cite{lu2023visual}&CTransPath/224&BioClinicalBert/512&0.456$|$0.461$|$\underline{0.755}&\underline{0.781}$|$\underline{0.723}$|$0.856&\underline{0.805}$|$0.782$|$0.881&0.802$|$0.792$|$0.866\\\tabucline[0.5pt]{-}
CONCH \cite{lu2023towards}&ViT-B/16-256&HistPathGPT/512&-&0.643$|$0.600$|$\underline{0.873}&0.796$|$\underline{0.797}$|$\underline{\textbf{0.961}}&\underline{0.807}$|$\underline{0.803}$|$\underline{0.915}\\\tabucline[0.5pt]{-}
CPLIP$_{1}$ (Ours)&ViT-B/16-224 (OoD)&GPT-2/77 (OoD)&\underline{0.502}$|$\underline{0.477}$|$0.705&0.500$|$0.544$|$0.722&0.754$|$0.749$|$0.865&0.761$|$0.788$|$0.821\\\tabucline[0.5pt]{-}
CPLIP$_{2}$ (Ours)&PLIP-ViT-B/32-224 (InD)&PLIP-GPT/347 (InD)&\textbf{0.591}$|$\textbf{0.587}$|$\textbf{0.827}&\textbf{0.824}$|$\textbf{0.786}$|$\textbf{0.889}&\textbf{0.844}$|$\textbf{0.855}$|$0.926&\textbf{0.854}$|$\textbf{0.835}$|$\textbf{0.936}\\\tabucline[1.5pt]{-}
Models (Merged Prompts)&Image encoder pretraining&Text encoder pretraining&CAM16&TCGA-BRCA&TCGA-RCC&TCGA-NSCLC\\\tabucline[1.5pt]{-}
CLIP baseline \cite{radford2021learning}&ViT-B/16-224&GPT-2/77&0.151$|$0.198$|$0.331&0.534$|$0.346$|$0.623&0.367$|$0.219$|$0.651&0.567$|$0.431$|$0.598\\\tabucline[0.5pt]{-}
BiomedCLIP \cite{zhang2023large}&ViT-B/16-224&PMB/256&0.337$|$0.402$|$0.564&0.532$|$0.441$|$0.837&0.807$|$0.773$|$0.903&0.777$|$0.761$|$0.861\\\tabucline[0.5pt]{-}
PLIP \cite{huang2023visual}&ViT-B/32-224&GPT/347&0.446$|$0.442$|$0.711&0.487$|$0.364$|$0.655&0.794$|$0.772$|$0.935&0.768$|$0.805$|$0.819\\\tabucline[0.5pt]{-}
MI-Zero \cite{lu2023visual}&CTransPath/224&BioClinicalBert/512&0.499$|$0.521$|$\underline{0.821}&0.833$|$0.821$|$0.905&0.871$|$0.855$|$0.933&0.881$|$0.871$|$0.944\\\tabucline[0.5pt]{-}
CONCH \cite{lu2023towards}&ViT-B/16-256&HistPathGPT/512&-&\underline{0.840}$|$\underline{0.839}$|$\underline{0.932}&\underline{0.893}$|$\underline{0.895}$|$\underline{0.973}&\underline{0.900}$|$\underline{0.900}$|$\underline{0.964}\\\tabucline[0.5pt]{-}
CPLIP$_{1}$ (Ours)&ViT-B/16-224 (OoD)&GPT-2/77 (OoD)&\underline{0.578}$|$\underline{0.551}$|$0.751&0.557$|$0.588$|$0.783&0.834$|$0.805$|$0.921&0.811$|$0.856$|$0.833\\\tabucline[0.5pt]{-}
CPLIP$_{2}$ (Ours)&PLIP-ViT-B/32-224 (InD)&PLIP-GPT/347 (InD)&\textbf{0.661}$|$\textbf{0.632}$|$\textbf{0.886}&\textbf{0.887}$|$\textbf{0.871}$|$\textbf{0.963}&\textbf{0.941}$|$\textbf{0.937}$|$\textbf{0.978}&\textbf{0.931}$|$\textbf{0.951}$|$\textbf{0.981}\\\tabucline[1.5pt]{-}
\end{tabu}
}}
\end{center}
\label{table9}
\end{table*}

\begin{table*}[t!]
\caption{Zero-shot segmentation performance comparison of gigapixel images in terms of dice score, precision, and recall with existing VL-based models in computational pathology on two independent datasets using the single prompt. OoD: Out-of-Domain and InD: In-Domain}
\begin{center}
\makebox[\linewidth]{
\scalebox{0.85}{
\begin{tabu}{|c|c|c|c|c|}
\tabucline[1.5pt]{-}
Models (Single prompts)&Image encoder pretraining&Text encoder pretraining&SICAP&DigestPath\\\tabucline[1.5pt]{-}
CLIP baseline \cite{radford2021learning}&ViT-B/16-224&GPT-2/77&0.367$|$0.599$|$0.605&0.367$|$0.492$|$0.511\\\tabucline[0.5pt]{-}
BiomedCLIP \cite{zhang2023large}&ViT-B/16-224&PMB/256&0.484$|$0.536$|$0.557&0.446$|$0.581$|$0.601\\\tabucline[0.5pt]{-}
PLIP \cite{huang2023visual}&ViT-B/32-224&GPT/347&0.549$|$0.605$|$0.644&0.426$|$0.526$|$0.541\\\tabucline[0.5pt]{-}
MI-Zero \cite{lu2023visual}&CTransPath/224&BioClinicalBert/512&0.587$|$0.651$|$0.726&0.599$|$0.648$|$0.691\\\tabucline[0.5pt]{-}
CONCH \cite{lu2023towards}&ViT-B/16-256&HistPathGPT/512&0.601$|$0.672$|$0.751&0.615$|$0.663$|$0.709\\\tabucline[0.5pt]{-}
CPLIP$_{1}$ (Ours)&ViT-B/16-224 (OoD)&GPT-2/77 (OoD)&0.591$|$0.661$|$0.681&0.491$|$0.581$|$0.602\\\tabucline[0.5pt]{-}
CPLIP$_{2}$ (Ours)&PLIP-ViT-B/32-224 (InD)&PLIP-GPT/347 (InD)&\underline{\textbf{0.654}}$|$\underline{0.704}$|$\underline{0.803}&\underline{0.685}$|$\underline{0.719}$|$\underline{0.754}\\\tabucline[0.5pt]{-}
CPLIP$_{3}$ (Ours)&CTransPath/224 (InD)&BioClinicalBert/512 (OoD)&0.633$|$0.702$|$0.791&0.665$|$0.711$|$0.744\\\tabucline[0.5pt]{-}
CPLIP$_{4}$ (Ours)&CTransPath/224 (InD)&PLIP-GPT/347 (InD)&\underline{0.651}$|$\textbf{0.715}$|$\textbf{0.806}&\textbf{0.687}$|$\textbf{0.722}$|$\textbf{0.761}\\\tabucline[1.5pt]{-}
\end{tabu}
}}
\end{center}
\label{table10}
\end{table*}

\begin{figure*}[t!]
\centering
\includegraphics[width=0.9\linewidth]{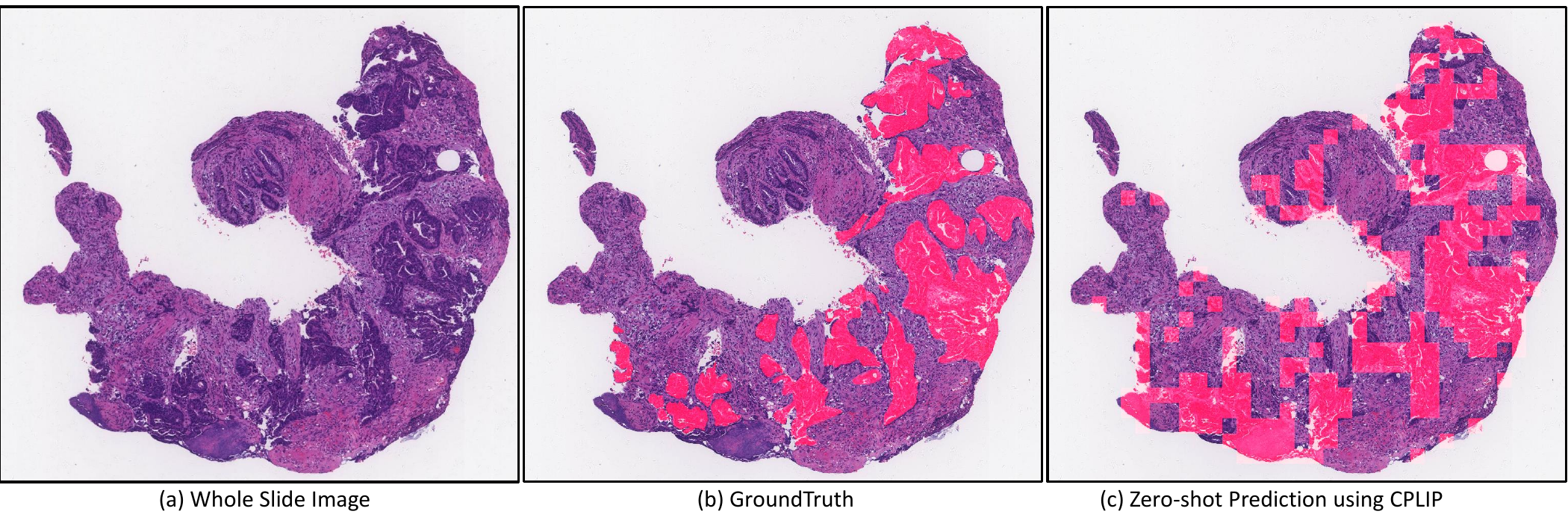}
\caption{Example of the segmentation results on one of the WSIs selected from DigestPath \cite{da2022digestpath} dataset using our proposed CPLIP algorithm.
Here, it is important to note that the segmentation task is posed as a tile-based zero-shot classification problem similar to CONCH \cite{lu2023towards}. 
The WSI is divided into tiles and the similarity scores for each tile are computed independently. 
However, instead of aggregating the scores across tiles into a single slide-level prediction, we map the tile-level scores to their corresponding spatial locations in the WSI and average the scores in the overlapping regions. 
Finally, each pixel is labeled with the class that has the top score, resulting in the formation of a detailed pixel-wise segmentation mask.}
\label{fig_seg}
\end{figure*}

\subsection{Tile-level Zero-shot Classification Results (Table \ref{table8})}
We performed zero-shot experiments on five different datasets for tile-level classification using only the testing split of each dataset. 
Table \ref{table8} shows the comparison with existing SOTA VL-based methods in terms of balanced accuracy, weighted average $F_{1}$, and AUROC scores using both single and merged prompts. 
Our proposed CPLIP algorithm achieved the best performance on all three metrics on all five datasets in both settings. 
CONCH obtained the second-best performance on the CRC100K and SICAP datasets, but its performance has not been reported on the remaining datasets.

Specifically, CPLIP obtained a 13.5$\%$ improvement in balanced accuracy, a 13.90\% improvement in weighted $F_{1}$, and a 2.10\% improvement in AUROC over CONCH on the CRC100K dataset using single prompts. 
Using merged prompts, CPLIP achieved a 3.20\% improvement in balanced accuracy, a 4.10\% improvement in weighted $F_{1}$, and a 0.10\% improvement in AUROC over CONCH on the CRC100K dataset. 
On the SICAP dataset, CPLIP achieved a 1.70\% improvement in balanced accuracy, and a 14.30\% improvement in weighted $F_{1}$ over CONCH using single prompts. 
Using merged prompts, CPLIP achieved an 8.70\% improvement in weighted $F_{1}$ over CONCH. 
For DigestPath and PanNuke datasets, CPLIP improved (0.6 $\%$, 3.2$\%$, 3.0$\%$) and (5.10$\%$, 5.20$\%$, 3.50$\%$) performance using merged prompts over the second best performers MI-Zero/PLIP. 
On the WSSS4LUAD dataset, CPLIP improved its performance over the second-best performer PLIP by 10.0\% in balanced accuracy, 9.10\% in weighted $F_{1}$, and 7.0\% in AUROC using merged prompts. 
\textit{Most of these performances are significantly better than the existing SOTA methods, demonstrating the advantages of our proposed CPLIP algorithm}.

\subsubsection{WSI-level Zero-shot Classification Results (Table \ref{table9})}
To extend zero-shot transfer to gigapixel WSIs, we used a method similar to MI-Zero \cite{lu2023visual}. 
For classification over $C$ classes, we first used the OTSU method to binarize the WSI into tissue and background regions. 
We then divided the tissue region into $N$ tiles, each of size $224 \times 224$ pixels. 
For each tile, we estimated an $\ell_{2}$-normalized embedding independently using the CPLIP image encoder. 
For each tile embedding, we computed cosine similarity scores with each text embedding, obtaining a set of $C$ similarity scores for each tile. 
To aggregate similarity scores across tiles, we used the top-$K$ pooling operator, averaging over the highest $K$ similarity scores for each class to obtain the slide-level similarity score. 
The class with the highest slide-level score was the predicted class. 
We chose $K \in {1, 5, 10, 50, 100}$ and reported the results for the $K$ with the highest balanced accuracy, weighted $F_{1}$, and AUROC scores for classification tasks.

In Table  \ref{table9}, we compared the zero-shot classification performance of our proposed CPLIP algorithm with existing SOTA VL-based computational pathology models on four independent datasets: CAM16, TCGA-BRCA, TCGA-RCC, and TCGA-NSCLC, using both single and merged prompts.
We also presented the performance of our proposed CPLIP algorithm in terms of different fine-tuned out-of-domain and in-domain image and text encoders.

CPLIP outperformed SOTA in-domain VL models, including PLIP, BiomedCLIP, MI-Zero, and CONCH, on all datasets, often by a significant margin. 
For example, in the case of lymph node metastasis classification in CAM16 using single and merged prompts, CPLIP$_{2}$ (in-domain) achieved zero-shot balanced accuracies of 59.10\% and 66.10\%, respectively, and outperformed the next best performing model, MI-Zero, by 13.50\% and 16.20\%.

For Non-Small Cell Lung Cancer (NSCLC) and Renal Cell Carcinoma (RCC) subtyping using a single prompt, our proposed CPLIP$_{2}$ (In-domain) model achieves zero-shot balanced accuracies of 85.40\% and 84.40\% respectively. 
This outperforms the next best models, CONCH and MI-Zero, by margins of 4.70\% and 5.20\% on NSCLC and 4.80\% and 3.90\% on RCC. 
With merged prompts, CPLIP$_{2}$ (In-domain) further improves to 93.10\% and 94.10\% balanced accuracy, exceeding CONCH by 3.10\% and 4.80\%. 
Similarly, on the more challenging invasive breast carcinoma (BRCA) subtyping task, our CPLIP$_{2}$ (In-domain) achieves 88.70\% zero-shot balanced accuracy, surpassing CONCH and MI-Zero by significant margins of 4.70\% and 5.40\%.
\textit{Overall, the proposed CPLIP$_{2}$ demonstrates SOTA performance on multiple cancer subtyping tasks using zero-shot learning.}

\subsubsection{Zero-shot Segmentation Results of Gigapixel Images (Table \ref{table10})}
We perform zero-shot segmentation of gigapixel WSIs similar to CONCH \cite{lu2023towards} using the same classification methods described above. 
We divide the WSI into tiles and compute similarity scores for each tile independently. 
However, instead of aggregating the scores across tiles into a single slide-level prediction, we map the tile-level scores to their corresponding spatial locations in the WSI and average the scores in overlapped regions. 
Finally, for each pixel, we assign the class with the highest score as the prediction, producing a pixel-level segmentation mask.

We used the official testing splits of the SICAP dataset (31 WSIs) for prostate tumor vs. normal tissue segmentation and DigestPath (250 large images) for colon malignant vs. benign tissue for zero-shot segmentation. 
Results are reported in Table \ref{table10} in terms of Dice score, precision, and recall to quantify the quality of the predicted segmentation mask relative to the ground truth using a single prompt during the inference stage.
Our proposed CPLIP algorithm outperforms other VL computational pathology models in both datasets. 
A visual result of the CPLIP algorithm is shown in Fig. \ref{fig_seg} using a sample WSI from the DigestPath dataset.

In SICAP, our best-performing CPLIP$_{2}$ and CPLIP$_{4}$ models achieve average Dice scores of 65.40\% and 65.10\%, respectively, outperforming CONCH (60.10\%), MI-Zero (58.70\%), PLIP (54.90\%), and BiomedCLIP (48.40\%) by a significant margin.
In DigestPath, our proposed best-performing in-domain CPLIP$_{4}$ and CPLIP$_{2}$ models achieved average Dice scores of 68.70\% and 68.50\%, respectively, outperforming CONCH (61.50\%), MI-Zero (59.90\%), PLIP (42.60\%), and BiomedCLIP (44.60\%) by a significant margin. 
\textit{Additionally, we found that despite the coarse-grained and zero-shot nature of the approach, CPLIP was able to produce reasonably accurate pixel-level segmentation masks, demonstrating the advantages of heterogeneous textual descriptions and histology images.}

\subsubsection{Computational Time Analysis}
We conducted our experiments on a DGX NVIDIA workstation with 256 GB of RAM and 4 Tesla V100 GPUs. 
At inference, our model only needs to first compute the image-text representation and then perform cosine similarity, which can be implemented efficiently using matrix multiplication. 
On the TCGA-BRCA dataset for cancer subtyping, CPLIP took an average of 3.2 minutes to process per WSI, depending on the value of $K$, while other VL models, including PLIP (2.90 minutes), MI-Zero (3.00 minutes), and BiomedCLIP (2.70 minutes), were faster. 
Overall, CPLIP is comparable in speed to other VL models at inference.


\clearpage
\onecolumn
\begin{longtable}{c|c}
\caption{Our proposed pathology prompts dictionary used during the construction of a bag of textual descriptions and a bag of visual concepts.}
\\
\hline
\toprule
\multicolumn{1}{l|}{\textbf{Alphabet}} & \multicolumn{1}{c}
{\textbf{Pathology Prompts Dictionary}}                           \\ \midrule
\multirow{10}{*}{\textbf{A}}           & Advanced breast Cancer; Antibody-Dependent cellular cytotoxicity;     Adenocarcinoma; Adenoma benign cancer;               \\ \cmidrule(l){2-2} 
                                       & Adenomatous polyp;   Adenocarcinoma of the lung; Atypical glandular cells;   Acinar pattern                                                         \\ \cmidrule(l){2-2} 
                                       & adenocarcinoma; Acinar growth pattern; Acinar predominant histological subtype;                                            \\ \cmidrule(l){2-2} 
                                       & Alanine aminotransferase / alanine transaminase; Anaplastic Large-cell Lymphoma; Acute lymphocytic leukemia;                                                        \\ \cmidrule(l){2-2} 
                                       & Adipose tissue/adipocytes; Acute myeloid leukemia;   Anaplastic; Alveolar rhabdomyosarcoma;                                \\ \cmidrule(l){2-2} 
                                       &  Alveolar Soft Part Sarcoma; Anaplastic Thyroid Cancer;                                                          \\ \midrule
\multirow{7}{*}{\textbf{B}}            & B-Cell Acute lymphoblastic leukemia; Breast cancer; Basal cell carcinoma;      B-cell lymphoma; Benign tissue;       \\ \cmidrule(l){2-2} 
                                       &Benign glands; Benign colon tissue;  Bladder cancer including melonoma; Benign rectal tissue;               \\ \cmidrule(l){2-2} 
                                       &   Benign essential blepharospasm; Bone marrow; BRCA1 and BRCA2; Brain Tumor;                          
                                           \\ \cmidrule(l){2-2} 
                                       & Breast invasive lobular carcinoma;  Benign multi-cystic peritoneal mesothelioma;  Breast invasive ductal carcinoma;                                              
                                                                        
                                                                              \\ \midrule
\multirow{11}{*}{\textbf{C}}           & Cancer; carcinoma; Cancer staging; Carcinoma in situ; Carotid body tumor;     Clear cell renal cell carcinoma;       \\ \cmidrule(l){2-2} 
                                       & Carcinoid tumor; 
 Carcinoma In Situ;   Chronic granulocytic leukemia; Cervical intraepithelial neoplasia;                   \\ \cmidrule(l){2-2} 
                                       & Chronic inflammatory bowel disease; Chromophobe renal cell carcinoma; Cirrhosis; Cell-mediated immunity;             \\ \cmidrule(l){2-2} 
                                       &  Chronic myeloid leukemia;  Chronic myelomonocytic leukemia; Cytomegalovirus; Comed—comedocarcinoma;                      \\ \cmidrule(l){2-2} 
                                       & Colectomy; Colitis; Colon polyp; Colonoscopy; Cancer-associated stroma;     Coloncancer adenocarcinoma debris;        \\ \cmidrule(l){2-2} 
                                       &  Core biopsy; 
 choroid plexus carcinoma;    Colorectal carcinoma/cancer; Colorectal adenocarcinoma;        \\ \cmidrule(l){2-2} 
                                       & Cerebrospinal fluid;   Circulating tumor cell; cancerous tissue; Cutaneous T-cell lymphoma; Cerebrovascular accident;                  \\ \cmidrule(l){2-2} 
                                       &  CC-Cervical cancer;    Colitis;                                     \\ \midrule
\multirow{2}{*}{\textbf{D}}            & Diffuse histolytic lymphoma; Distant recurrence; Debris or dead cell;               \\ \cmidrule(l){2-2} 
                                       & Diffuse large B-cell lymphoma; Dukes staging system; Dysplasia;                     \\ \midrule
\multirow{3}{*}{\textbf{E}}            & Early-Stage invasive breast cancer; Epstein-barr virus; Esophageal cancer;  Epidermal growth Factor Receptor;    \\ \cmidrule(l){2-2} 
                                       &  Extra skeletal myxoid chondrosarcoma;  Estrogen receptor;                                                            \\ \midrule
\multirow{2}{*}{\textbf{F}}            & Formalin fixed paraffin embedded tissues; Fluorescence In Situ hybridization;    Fibrosis;                                                                         \\ \midrule
\multirow{2}{*}{\textbf{G}}            & Gastrointestinal stromal tumors; Gastrointestinal cancer; Gleason score;  Gleason score;                                                                             \\ \midrule
\multirow{5}{*}{\textbf{H}}            & Hand foot syndrome; Hepatocellular carcinoma; Hairy cell leukemia; Hodgkin’s disease;   Hyperplasia;              \\ \cmidrule(l){2-2} 
                                       &  Human epidermal growth factor receptor 2;      Hereditary nonpolyposis colon cancer;                           \\ \cmidrule(l){2-2} 

                                         &    Human immunodeficiency virus; Hereditary nonpolyposis colon cancer;    Human T-cell Leukemia;                      \\ \cmidrule(l){2-2}
                                         
                                       & Head and neck squamous cell carcinoma; Human papillomavirus;  Hidradenoma papiliferum;                             \\ \midrule
\multirow{4}{*}{\textbf{I}}            & Immunohistochemistry; Invasive cancer; In Situ hybridization;        Invasive ductal carcinoma of the breast;               \\ \cmidrule(l){2-2} 

& IInvasive carcinoma of the breast;   Invasive lobular carcinoma; Inflammatory cells;             \\ \cmidrule(l){2-2}
                                       &    Immune cells;  Inflammatory bowel diases;                                                            \\ \midrule
                                       
\multirow{4}{*}{\textbf{L}}            & Langerhan’s cell histiocytosis; lentigo maligna melanoma; Lung cancer; Lung squamous cell carcinoma;              \\ \cmidrule(l){2-2} 
                                       & Lung adenocarcinoma; Lymph Node;  Leipidic pattern adenocarcinoma; Lymphoid infiltrate;                                   \\ \cmidrule(l){2-2} 
                                       & Leipidic predominant histologuical subtype; Lymphocytes; Liver cancer;              \\ \midrule
\multirow{8}{*}{\textbf{M}}            & Malignant; Mouth and throat cancer; Mastectomy; Myelodysplastic syndromes; Multiple endocrine neoplasia;         \\ \cmidrule(l){2-2} 
                                       &  Malignant colon tissue; Malignant rectal tissue;  Mucus/Mucin; Metastasis; Muscularis propria; Mucinous carcinoma;                      \\ \cmidrule(l){2-2} 
                                       & Micropappillary pattern; Micropappillary pattern adenocarcinoma; Micropappillary growth pattern;                   \\ \cmidrule(l){2-2} 
                                       &  Micropappillary predominant histological subtype; Muscularis mucosa; Metastatic cells; Malignant melanoma;                 \\ \cmidrule(l){2-2} 
                                       & Malignant peripheral nerve Sheath tumor; Malignant rhabdoid tumor;       Microsatellite instability; Microsatellite stability;              
                                                                  \\ \midrule
\multirow{4}{*}{\textbf{N}}            & Normal adjacent tissue; Nevoid basal cell carcinoma Syndrome; Non-Hodgkin’s lymphoma;                       \\ \cmidrule(l){2-2} 
                                       &  Nodular melanoma; Non melanoma skin ccancer;  Nasopharyngeal cancer; Node-negative breast cancer;              \\ \cmidrule(l){2-2} 
                                       &   Node-Positive breast Cancer;  Non-Small cell lung cancer; Necrosis; Neoplasi; Neoplasm; Neutrophils;           \\ \midrule
\textbf{O}                             & Osteogenic Sarcoma; Ovarian cancer;                                                 \\ \midrule
\multirow{13}{*}{\textbf{P}}           & Pathologic (or Histologic) grade well differentiated;  Pathologic (or Histologic) Grade moderately differentiated;                                       \\ \cmidrule(l){2-2} 
                                       & Pathologic (or Histologic) grade poorly differentiated; pathologic (or Histologic) grade undifferentiated;                                          \\ \cmidrule(l){2-2} 
                                       & Pathologic stage; Peripheral blood mononuclear cells; Parkinson’s disease; Primary lymphoma of bone; Polyp;          \\ \cmidrule(l){2-2} 
                                       &  Progesterone receptor; Prostate-Specific antigen; Prostrate cancer; Prostrate cancer with gleason grade 3;                                                         \\ \cmidrule(l){2-2} 
                                       & Prostrate cancer with Gleason grade 4;    Prostrate cancer with gleason grade 5;  Prostrate adenocarcinoma;                                                                   \\ \cmidrule(l){2-2} 
                                       &  Prostatic adenocarcinoma; Papillary renal cell carcinoma; Papillary pattern adenocarcinoma;\\ \cmidrule(l){2-2} 
                                       &  Pancreatic cancer; Papillary growth pattern; Papillary tumor; Papilloma;                                                                                                                                        \\ \midrule
\multirow{2}{*}{\textbf{R}}            & Renal cell carcinoma; Renal cell carcinoma of chromophore type; Rhabdomyosarcoma;                                                                                   \\ \midrule
\multirow{4}{*}{\textbf{S}}            & Sarcoma; Squamous Cell Carcinoma; Small Cell Lung Cancer; Secondary Score;          \\ \cmidrule(l){2-2} 
                                       & Synchronous cancer; Solid pattern adenocarcinoma; Solid growth pattern;  Smooth muscle;            \\ \cmidrule(l){2-2} 
                                       & Stromal tissue; Stromal cells; Skin cancer; Stroma associated tumor; Sialadenoma papilliferum;                                           \\ \midrule
\multirow{3}{*}{\textbf{T}}            & Transitional cell carcinoma; Thrombocytopenia; classification of malignant tumors;  \\ \cmidrule(l){2-2} 
                                       & Tumor; Tumor grade; Tumor-associated stroma; Tumor infiltrating lymphocytes;        \\ \cmidrule(l){2-2} 
                                       & Tumor epithelial tissue; Testicular cancer;                                         \\ \midrule
\multirow{2}{*}{\textbf{U}}            & Ulcerative colitis; Urinary bladder Cancer; Urinary bladder adenocarcinoma;    Urinary bladder tissue;                                                                 \\                                                                              \\ \midrule
\textbf{W}                             & White Blood cell count; Waldenstrom’s macroglobulinemia;                                                                                                    \\ \midrule
\textbf{Y}                             & Yolk sac Tumor;                                                                          \\ \bottomrule
\label{table_dictionary}
\end{longtable}





\end{document}